\documentclass[10pt,twocolumn,letterpaper]{article}

\usepackage[pagenumbers]{cvpr} 

\usepackage[linesnumbered,ruled,vlined]{algorithm2e}
\usepackage{amsmath}
\usepackage{amsfonts}
\usepackage{bm}
\usepackage{booktabs}
\usepackage{multirow}
\usepackage{threeparttable}
\usepackage{booktabs}
\usepackage{siunitx}
\usepackage[margin=1in]{geometry} 
\usepackage{multirow} 
\usepackage{graphicx}
\usepackage{capt-of}
\usepackage{caption}
\usepackage{subcaption}

\definecolor{cvprblue}{rgb}{0.21,0.49,0.74}
\usepackage[pagebackref,breaklinks,colorlinks,allcolors=cvprblue]{hyperref}

\def\paperID{*****} 
\def\confName{CVPR}
\def\confYear{2026}

\title{ForgeDreamer: Industrial Text-to-3D Generation with Multi-Expert LoRA and Cross-View Hypergraph}

\author{
Junhao Cai\textsuperscript{1,}\thanks{Equal contribution. \quad $^\dagger$Corresponding author.} \quad 
Deyu Zeng\textsuperscript{1, 2,}\footnotemark[1] \quad 
Junhao Pang\textsuperscript{1} \quad 
Lini Li\textsuperscript{1} \quad 
Xiaopin Zhong\textsuperscript{1,$\dagger$} \quad 
Zongze Wu\textsuperscript{1}\\
[2mm] 
\textsuperscript{1}Shenzhen University, Shenzhen, Guangdong 518060, China\\
\textsuperscript{2}Guangzhou Maritime University, Guangzhou, Guangdong 510725, China\\
[1mm]
{\tt\small caijunhao27@gmail.com} \quad {\tt\small zengdeyu@gzmtu.edu.cn} \\
{\tt\small \{2500092013, 2500092003\}@mails.szu.edu.cn} \quad {\tt\small \{xzhong, zzwu\}@szu.edu.cn}
\vspace{-5mm}
}

\begin{document}
\maketitle
\begin{abstract}
Current text-to-3D generation methods excel in natural scenes but struggle with industrial applications due to two critical limitations: domain adaptation challenges where conventional LoRA fusion causes knowledge interference across categories, and geometric reasoning deficiencies where pairwise consistency constraints fail to capture higher-order structural dependencies essential for precision manufacturing. 
We propose a novel framework named ForgeDreamer addressing both challenges through two key innovations. First, we introduce a Multi-Expert LoRA Ensemble mechanism that consolidates multiple category-specific LoRA models into a unified representation, achieving superior cross-category generalization while eliminating knowledge interference. Second, building on enhanced semantic understanding, we develop a Cross-View Hypergraph Geometric Enhancement approach that captures structural dependencies spanning multiple viewpoints simultaneously.
These components work synergistically improved semantic understanding, enables more effective geometric reasoning, while hypergraph modeling ensures manufacturing-level consistency. Extensive experiments on a custom industrial dataset demonstrate superior semantic generalization and enhanced geometric fidelity compared to state-of-the-art approaches. Code is available at \url{https://github.com/Junhaocai27/ForgeDreamer}
\end{abstract}
\section{Introduction}
\label{sec:intro}

Text-to-3D generation has emerged as a transformative technology, enabling the creation of diverse 3D content from natural language descriptions. Recent breakthroughs, led by pioneering works such as DreamFusion \cite{poole2022dreamfusion} and subsequent advances like ProlificDreamer \cite{wang2023prolificdreamer} and LucidDreamer \cite{liang2024luciddreamer}, have demonstrated remarkable capabilities in generating imaginative 3D assets for creative applications. These methods typically leverage Score Distillation Sampling (SDS) \cite{lin2023magic3d, chen2023fantasia3d, metzer2023latent} to distill knowledge from pretrained 2D diffusion models \cite{rombach2022high, ho2020denoising, ramesh2022hierarchical, luo2024skipdiff} into 3D representations \cite{wang2021nerf, mildenhall2021nerf, wu2024recent, muller2022instant}, achieving impressive results for natural scene generation.

\begin{figure}[t!]
\centering
\includegraphics[width=0.8\columnwidth]{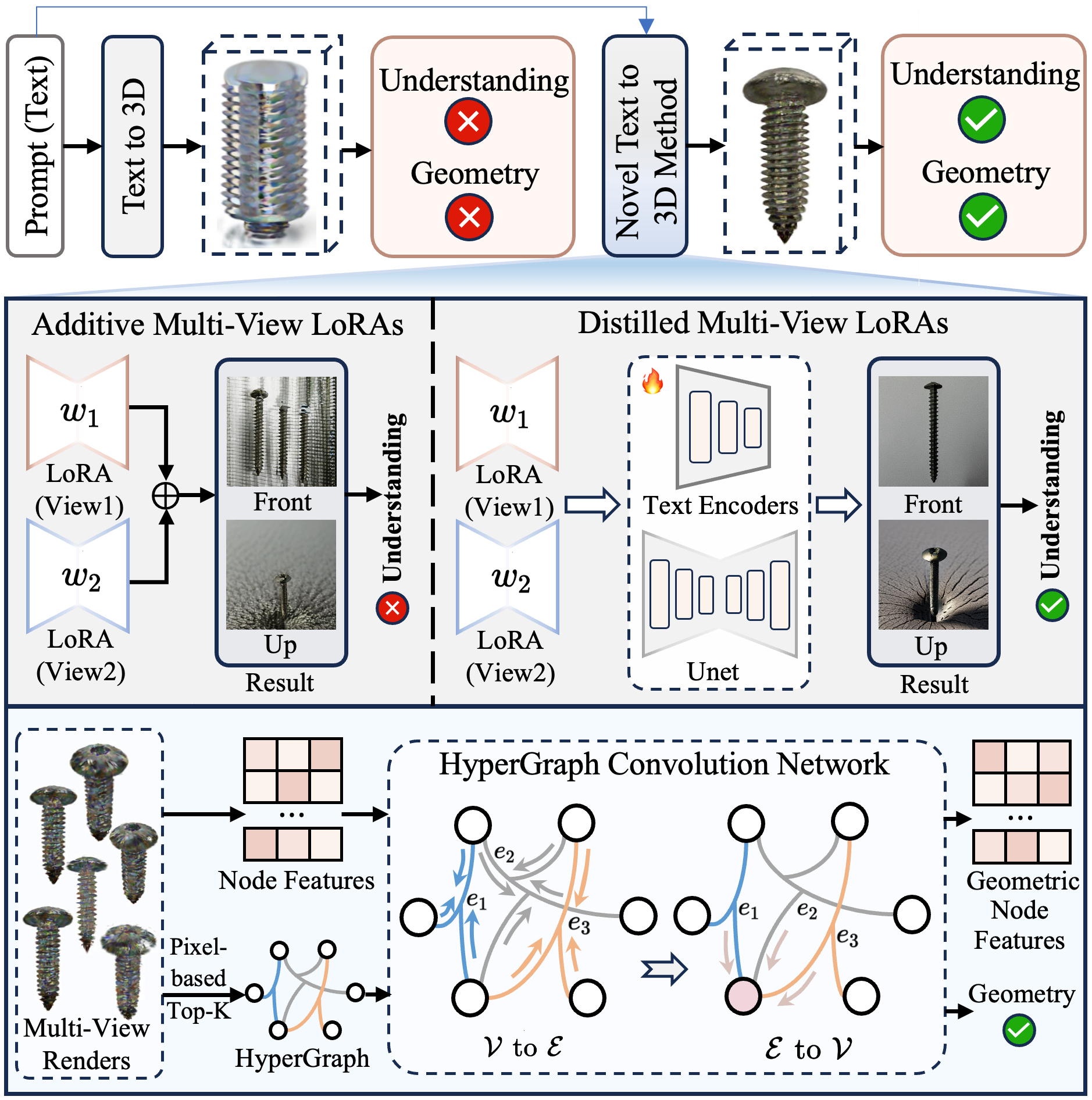} 
\caption{Overall Framework Architecture: Multi-Expert LoRA Ensemble Framework and Cross-View Hypergraph Enhancement}
\label{fig1}
\vspace{-5mm}
\end{figure}


\begin{figure}[h]
\centering
\includegraphics[width=1.0\columnwidth]{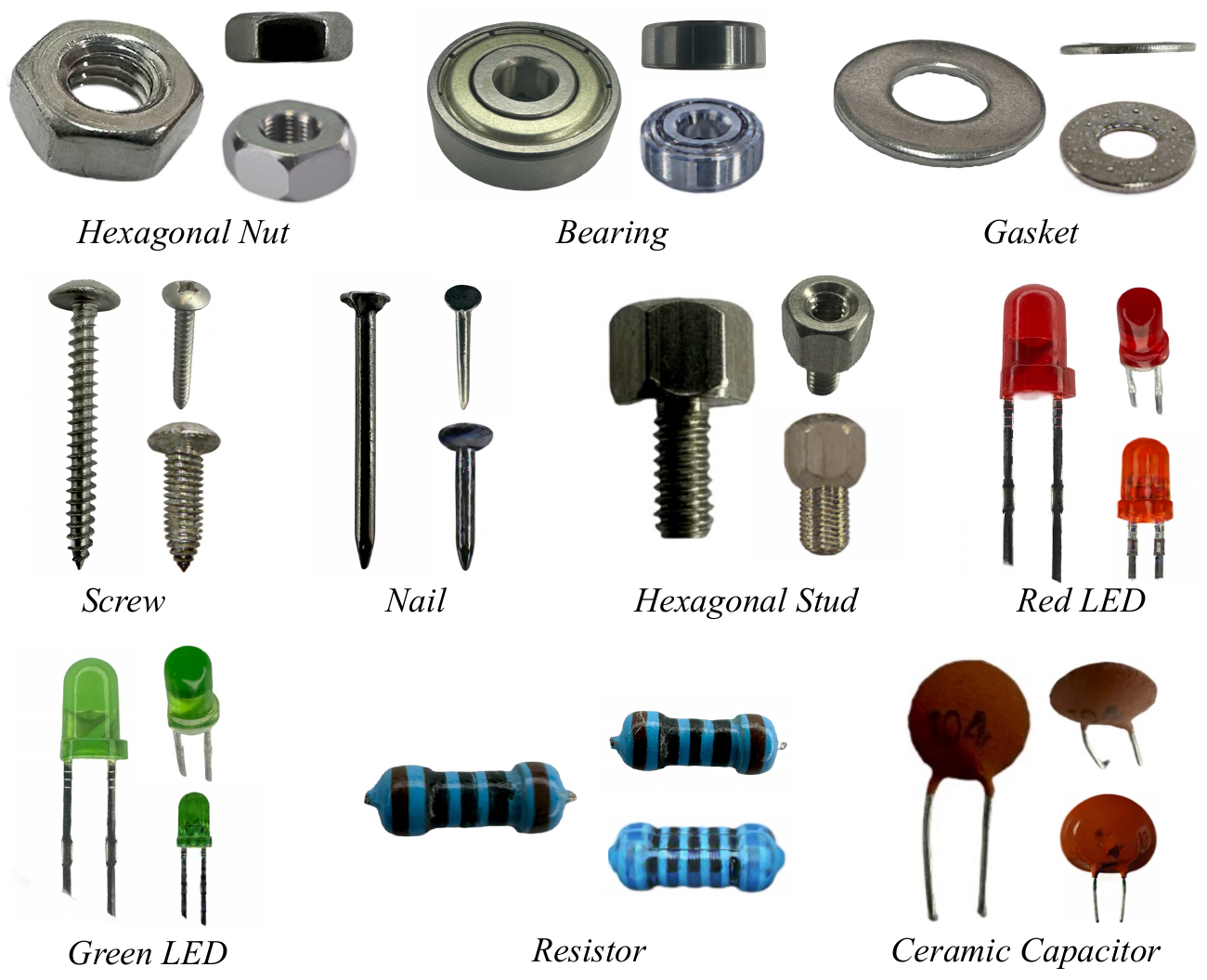} 
\caption{Overview of custom-built multi-view industrial dataset and 3D generation results. The bottom-right image presents the corresponding 3D generation result produced by our method.}
\label{fig2}
\vspace{-5mm}
\end{figure}

The field has witnessed rapid progress with diverse methodological innovations \cite{liu2023zero, shi2023mvdream, hong2023lrm, luo2021diffusion, zhou20213d, luo2022lattice}. Era3D \cite{li2024era3d} introduced efficient row-wise attention for high-resolution multiview diffusion, while RichDreamer \cite{qiu2024richdreamer} enhanced detail richness through normal-depth diffusion models. Point-E \cite{nichol2022point} pioneered point cloud generation from complex prompts, and recent works like PI3D \cite{liu2024pi3d} and DreamPropeller \cite{zhou2024dreampropeller} have further improved generation efficiency through pseudo-image diffusion and parallel sampling strategies. Additionally, consistency-focused approaches such as Consistent3D \cite{wu2024consistent3d} have addressed deterministic sampling challenges, while subject-driven methods like DreamBooth3D \cite{raj2023dreambooth3d} have enabled personalized 3D content creation.

However, despite significant progress in general-purpose text-to-3D generation, applying these techniques to industrial applications reveals critical limitations \cite{willis2022joinable, lambourne2021brepnet, zuo2024reconstruction}. Industrial components like mechanical fasteners, electronic parts, and precision-machined surfaces demand both semantic accuracy and geometric precision that current approaches struggle to deliver. Existing methods face two fundamental challenges: (1) Domain Gap: models trained on natural scenes poorly understand industrial-specific semantics and geometric requirements; (2) Geometric Reasoning: traditional pairwise consistency constraints inadequately capture the complex structural dependencies essential for industrial precision, leading to artifacts in threading patterns, connector interfaces, and dimensional accuracy.

While Low-Rank Adaptation (LoRA) \cite{hu2022lora, dettmers2023qlora, zhang2023adalora} offers a parameter-efficient approach for domain adaptation, conventional LoRA fusion strategies suffer from knowledge interference when combining multiple category-specific adaptations. Recent works have explored LoRA composition for image generation \cite{zhong2024multi, kumari2023multi} and domain bridging applications such as X-Dreamer \cite{macreating}, but the challenges of cross-category generalization in 3D domains remain largely unaddressed. Control-based approaches like DreamControl \cite{huang2024dreamcontrol} have shown promise in enhancing text-to-3D generation with 3D self-priors, yet they still operate under domain-specific constraints that limit industrial applicability. Similarly, existing geometric consistency formulations \cite{yu2021pixelnerf, chen2021mvsnerf, wang2021ibrnet}, including the advanced Interval Score Matching approach \cite{liang2024luciddreamer}, operate under pairwise assumptions \cite{niemeyer2021giraffe, schwarz2020graf} that fail to capture the higher-order structural relationships \cite{groueix2018papier, genova2020local} crucial for industrial precision.

To address these challenges, we propose a systematic framework called ForgeDreamer for high-fidelity industrial text-to-3D generation that tackles the two fundamental limitations, as illustrated in \Cref{fig1}:
(1) We address the semantic understanding challenge through a Multi-Expert LoRA Ensemble Framework that employs teacher-student knowledge distillation to consolidate multiple category-specific LoRA models into a unified representation, which achieves superior cross-category generalization for industrial domains.
(2) Building on enhanced semantic understanding, we tackle the geometric precision challenge through Cross-View Hypergraph Geometric Enhancement. Drawing inspiration from hypergraph neural networks \cite{feng2019hypergraph, gao2022hgnn+}, we formulate geometric consistency as a hypergraph learning problem that captures higher-order structural dependencies across multiple viewpoints simultaneously, moving beyond the limitations of pairwise consistency assumptions.

Moreover, existing industrial datasets such as MVTec 3D-AD \cite{bergmann2021mvtec} and Real-IAD\cite{wang2024real, zuo2025padiff} prove unsuitable for text-to-3D generation due to limited viewpoints and inconsistent 
imaging conditions (\textit{see Appendix for experimental validation}). Therefore, we construct 
a controlled multi-view dataset for reliable text-to-3D industrial generation as shown
 in \Cref{fig2}.

Our framework operates on 3D Gaussian Splatting \cite{kerbl20233d} for its superior efficiency in high-resolution rendering, essential for capturing intricate industrial details. 
Our main contributions are summarized as follows:
\begin{itemize}
    \item We propose a novel teacher-student distillation framework that effectively consolidates multiple category-specific LoRA models into a unified representation, achieving superior cross-category generalization for industrial domains while avoiding knowledge interference.
    \item We introduce a Cross-View Hypergraph Enhanced Higher-Order Geometric Gradient Loss that captures structural dependencies spanning multiple viewpoints, addressing the fundamental limitations of pairwise geometric consistency assumptions.
    \item We conduct extensive experiments on a custom-built multi-view industrial dataset, demonstrating superior performance in both generation quality and semantic generalization compared to existing state-of-the-art methods.
\end{itemize}
\section{Related Work}
\label{sec:rel_work}

\subsection{Text-to-3D Generation} 
DreamFusion \cite{poole2022dreamfusion} pioneered text-to-3D generation through Score Distillation Sampling (SDS), which optimizes a 3D representation $\theta$ by minimizing:
\begin{align}
    \mathcal{L}_{\text{SDS}}(\theta) = \mathbb{E}_{t,\boldsymbol{\epsilon},c}[\omega(t)\|\boldsymbol{\epsilon}(\bm{x}_t,t,y)-\boldsymbol{\epsilon}\|_2^2]
\end{align}
where $\bm{x}_t = \sqrt{\bar{a}_t}\bm{x}_0+\sqrt{1-\bar{a}_t}\boldsymbol{\epsilon}$ represents the noisy latent, and $\bm{x}_0=g(\theta,c)$ is the rendered view. However, as analyzed in LucidDreamer \cite{liang2024luciddreamer}, SDS suffers from over-smoothing due to inconsistent pseudo-ground-truths $\hat{\bm{x}}^t_0$. Recent works like ProlificDreamer \cite{wang2023prolificdreamer} and LucidDreamer \cite{liang2024luciddreamer} have addressed this through Variational Score Distillation and Interval Score Matching respectively, but primarily focus on natural scenes rather than industrial domains.

\begin{figure*}[h!]
\centering
\includegraphics[width=1.0\textwidth]{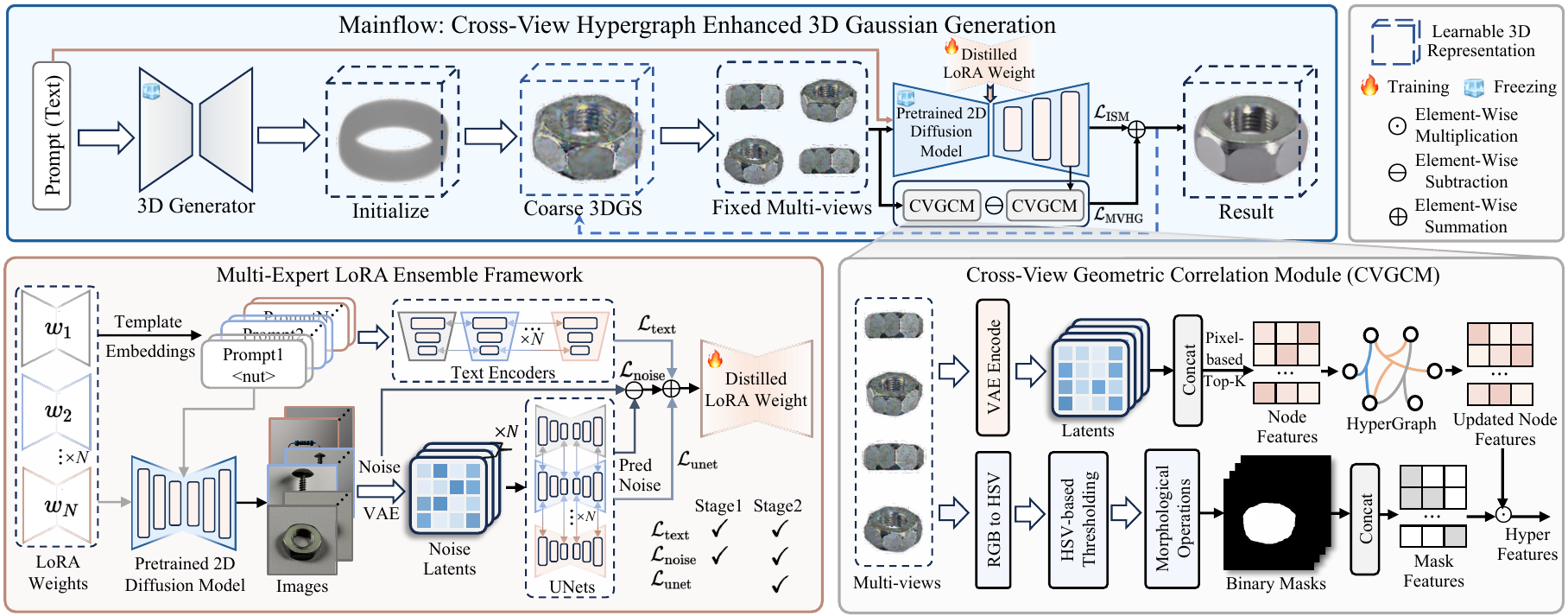} 
\caption{Architecture of our industrial Text-to-3D generation framework - ForgeDreamer. Top: Cross-view Hypergraph Enhanced 3D Gaussian Generation pipeline with Cross-View Geometric Correlation Module (CVGCM). Bottom: Multi-Expert LoRA Ensemble Framework for cross-category industrial knowledge integration.}
\label{fig3}
\end{figure*}

\subsection{Domain Adaptation Challenge} 
Current text-to-3D models face a fundamental limitation when applied to industrial domains: semantic gap between natural scene training data and industrial object requirements. 

Low-Rank Adaptation (LoRA) \cite{hu2022lora} enables efficient fine-tuning by decomposing weight updates as $\bm{W} = \bm{W}_0 + \bm{B}\bm{A}$, where $\bm{B}\in\mathbb{R}^{d\times r}$ and $\bm{A}\in \mathbb{R}^{r\times k}$ with rank $r\ll \text{min}(d,k)$. 

While individual LoRA adapters can capture category-specific knowledge, they suffer from knowledge isolation and lack cross-category generalization capabilities essential for diverse industrial applications.

Traditional LoRA fusion approaches employ simple additive combination,
$    \bm{W}_{\text{combined}} = \bm{W}_{\text{base}}+\sum_{i} \bm{W}^{(i)}_{\text{LoRA}}$.
However, this naive fusion leads to knowledge interference where conflicting category-specific features degrade overall performance, as evidenced by the declining cosine similarity scores in \Cref{tab:concept_preservation_compressed} as the number of LoRAs increases.

\subsection{Multi-View Geometric Consistency} Existing text-to-3D methods rely on view-independent geometric constraints, treating consistency as pairwise optimization problems. LucidDreamer's Interval Score Matching formulates consistency through interval-based score matching:
\begin{align}
    \mathcal{L}_{\text{ISM}}(\theta) = \mathbb{E}_{t,c}[\omega(t)\|\boldsymbol{\epsilon}_{\phi}(\bm{x}_t,t,y)-\boldsymbol{\epsilon}(\bm{x}_s,s,\Phi)\|_2^2]
\end{align}
where $s = t-\delta_T$. While effective, this approach still operates under pairwise assumptions and lacks the higher-order structural reasoning essential for complex industrial geometries with precise manufacturing requirements.

\subsection{Industrial 3D Generation Challenges} 
Current methods face fundamental limitations when applied to industrial domains: (1) semantic gap between natural scene training data and industrial object requirements, and (2) limited geometric reasoning for precision-critical applications. Our work addresses these challenges through multi-expert knowledge consolidation and hypergraph-based geometric modeling that captures structural dependencies beyond pairwise relationships.
\section{Methodology}
\label{sec:methodology}

Industrial text-to-3D generation faces two fundamental challenges that existing methods struggle to address effectively. First,  direct application of pre-trained diffusion models to industrial domains suffers from poor semantic understanding and limited cross-category generalization, as these models are primarily trained on natural scene datasets. Second, existing models exhibit limited geometric reasoning capabilities, particularly in capturing higher-order structural dependencies essential for precise industrial geometries.
\subsection{Overview}
\Cref{fig3} illustrates our systematic approach, named ForgeDreamer, to industrial text-to-3D generation, following a clear progression from semantic enhancement to geometric refinement to unified optimization: (1) Our Multi-Expert LoRA Ensemble Framework consolidates multiple category-specific LoRA models through teacher-student distillation, producing a unified representation that understands industrial component categories without knowledge interference. (2) Building on enhanced semantic understanding, our Cross-View Hypergraph Geometric Enhancement formulates consistency as a hypergraph learning problem, capturing higher-order structural dependencies across multiple viewpoints simultaneously. (3) We integrate both enhancements in an industrial-optimized pipeline where distilled LoRA weights provide semantic guidance while hypergraph modeling ensures geometric precision, together enabling high-fidelity industrial 3D generation. Overall, the key insight is that improving semantic understanding enables more effective geometric modeling, as the hypergraph operates on semantically meaningful features rather than poorly understood representations.

\subsection{Multi-Expert LoRA Ensemble Framework}
Industrial text-to-3D generation first requires robust semantic understanding of technical components. However, existing diffusion models trained on natural scenes lack domain-specific knowledge for industrial objects. While individual LoRA adapters can capture category-specific features, they suffer from knowledge isolation and interference when combined.

To address these limitations, we propose a teacher-student distillation framework that consolidates multiple category-specific LoRA experts into a unified representation, enabling cross-category generalization while preserving domain expertise. 

To construct this robust teacher-student architecture, designed to capture intricate geometric details for industrial applications, we begin with a set of pre-trained LoRA weights $\{\mathcal{L}_i\}^N_{i=1}$ and their corresponding trigger words $\{w_i\}^N_{i=1}$. Each teacher model $T_i$ is formed by loading LoRA weights $\mathcal{L}_i$ into the base Stable Diffusion model:
\begin{align}
    T_i = \text{SD}_{\text{base}} \oplus \mathcal{L}_i
\end{align}
where $\oplus$ denotes the LoRA integration operation. Each teacher $T_i$ specializes in generating high-quality samples for its corresponding trigger word $w_i$.

For each teacher $T_i$, we generate diverse 2D samples by incorporating multi-view prompts:
$$\begin{aligned}
    P^{view}_i = '\text{A photo of a} \ \{\text{screw\_front}\}/\{\text{screw\_up}\}'  
\end{aligned}$$

The generated images are encoded into latent space via VAE encoder: $z^{(i)} = VAE_{enc}(x^{(i)})$, forming individual datasets $D_i = {(z^{(i)}, P^{view}_i)}$ for each teacher model. The complete distillation procedure is detailed in \Cref{alg:multi_lora_distill_vlined}.

\begin{algorithm}[b]
\KwIn{LoRA weights $\{\mathcal{L}_i\}_{i=1}^N$, triggers $\{w_i\}_{i=1}^N$}
\KwOut{Distilled student model $S$}
\BlankLine
\For{each $\mathcal{L}_i$, $w_i$}{
    Load $\mathcal{L}_i$ into teacher $T_i$, sample with $w_i$ \\
    $\mathcal{D}_i$: VAE-encoded latent–text pairs\\
}
Initialize student model $S$\\
\BlankLine
\textbf{Stage 1: Train Text Encoder}\\
\For{each $(T_i, (\bm{x}_{\text{latent}}, w_i))$ from datasets $\{\mathcal{D}_i\}$}{
    $\mathcal{L}_{\text{text}} = \sum_l \alpha_l \cdot \text{MSE}(\text{Pool}(\bm{f}_T^l), \text{Pool}(\bm{f}_S^l))$ \tcp*{$\bm{f}$: text encoder features}
    $\mathcal{L}_{\text{noise}} = \text{MSE}(\hat{\boldsymbol{\epsilon}}_S, \boldsymbol{\epsilon})$\\
    Update $S_{\text{text}}$ with $\lambda_1^{\text{adp}} \mathcal{L}_{\text{text}} + \lambda_2^{\text{adp}} \mathcal{L}_{\text{noise}}$\\
}
\BlankLine
\textbf{Stage 2: Train Text Encoder \& UNet}\\
\For{each $(T_i, (\bm{x}_{\text{latent}}, w_i))$ from datasets $\{\mathcal{D}_i\}$}{
    $\mathcal{L}_{\text{text}} = \sum_l \alpha_l \cdot \text{MSE}(\text{Pool}(\bm{f}_T^l), \text{Pool}(\bm{f}_S^l))$\\
    $\mathcal{L}_{\text{unet}} = \sum_m \beta_m \cdot \text{MSE}(\bm{u}_T^m, \bm{u}_S^m)$ \tcp*{aligned UNet features}
    $\mathcal{L}_{\text{noise}} = \text{MSE}(\hat{\boldsymbol{\epsilon}}_S, \boldsymbol{\epsilon})$\\
    Update $S$ with $\gamma_1^{\text{adp}} \mathcal{L}_{\text{text}} + \gamma_2^{\text{adp}} \mathcal{L}_{\text{unet}} + \gamma_3^{\text{adp}} \mathcal{L}_{\text{noise}}$\\
}
\BlankLine
\KwRet{$S$}
\caption{Multi-LoRA Distillation}
\label{alg:multi_lora_distill_vlined}
\end{algorithm}

In the first stage, we exclusively train the student model's text encoder $S_{text}$ while keeping the UNet frozen. This design prevents catastrophic forgetting and ensures stable feature learning. The second stage simultaneously optimizes both text encoder and UNet using an alternating training paradigm that switches between noise prediction and feature alignment at fixed intervals.

During training, we employ a round-robin strategy to ensure balanced knowledge transfer from all teachers. Unlike additive fusion, our distillation approach learns to resolve conflicts by:
\begin{itemize}
    \item The MSE-based alignment losses encourage the student to find a common feature space that accommodates all teacher expertise.
    \item The two-stage approach allows the model to first establish a stable semantic foundation before integrating complex geometric reasoning.
\end{itemize}
By incorporating view-specific training data, the distilled model learns to maintain semantic consistency across different perspectives, crucial for industrial 3D generation where viewpoint-dependent features are essential.

\subsection{Cross-view Hypergraph Enhanced Higher-Order Geometric Gradient Loss}
While our distilled LoRA provides enhanced semantic understanding, industrial 3D generation demands geometric precision that goes beyond semantic accuracy. Traditional pairwise consistency constraints fail to capture the higher-order structural dependencies essential for precision manufacturing.

Thus, we formulate geometric consistency as a hypergraph learning problem, capturing structural dependencies across multiple viewpoints simultaneously rather than treating them as independent pairwise constraints.

\subsubsection{Hypergraph Geometric Modeling}
Unlike direct operations on point clouds \cite{di2025hyper}, we first propose a cross-view hypergraph formulation that fundamentally shifts from pairwise to higher-order geometric reasoning. Given multi-view latent representations $\bm{Z} = \{ \bm{z}^{(i)}\in\mathbb{R}^{H\times W \times C} \}^N_{i=1}$,
 we first construct a unified node feature representation by treating each pixel as an individual node.

For each view $i$, we reshape the latent representation $\bm{z}^{(i)} \in \mathbb{R}^{H\times W \times C}$ into a flattened format and concatenate across all views to form a comprehensive node feature matrix:
\begin{align}
    \bm{F} = \text{Concat}([\bm{z}^{(1)}_{\text{flat}},\bm{z}^{(2)}_{\text{flat}},\ldots,\bm{z}^{(N)}_{\text{flat}}])\in\mathbb{R}^{(N\cdot H\cdot W)\times C}
\end{align}
where $\bm{z}^{(i)}_{\text{flat}} \in \mathbb{R}^{(H \cdot W) \times C}$ represents the flattened latent features from view $i$. This results in a node feature matrix $\bm{F} \in \mathbb{R}^{(4 \cdot 64 \cdot 64) \times 4}$ for our experimental setup with 4 views and 4-dimensional latent channels.

Unlike conventional approaches that rely on spatial correspondence, we construct hypergraph $\mathcal{H} = (\mathcal{V}, \mathcal{E})$ through feature-based similarity relationships. For each node $v_i$ represented by feature vector $\mathbf{f}_i$ (the $i$-th row of $\bm{F}$), we establish hyperedges by computing feature similarities with all other nodes:
\begin{align}
    e_i = \{ v_j : v_j \in \text{TopK}(\text{sim}(\bm{f}_i,\bm{f}_j),k) \}
\end{align}
where $\text{sim}(\cdot, \cdot)$ computes the cosine similarity between feature vectors, and $\text{TopK}(\cdot, k)$ selects the $k$ most similar nodes to form a hyperedge. This feature-driven hyperedge construction enables the model to discover semantic and geometric correspondences across multiple viewpoints that may not align spatially but share similar structural characteristics.
\subsubsection{Cross-View Geometric Consistency Through Hypergraph Neural Networks}
We formulate geometric consistency as a hypergraph learning problem. Hypergraph Neural Networks (HGNN) is employed to aggregate cross-view geometric information through iterative message passing:
\begin{align}
    \bm{h}_v^{(l+1)}=\sigma(\bm{W}^{(l)} \sum_{e\in \mathcal{E}(v)}  \frac{1}{|\mathcal{E}(v)|} \text{AGG} (\{\bm{h}_u^{(l)}:u\in e\}))
\end{align}
where $\bm{h}_v^{(l)}$ represents the feature vector of node $v$ at layer $l$, $\mathcal{E}(v)$ denotes the set of hyperedges incident to node $v$, and uniform weighting $\frac{1}{|\mathcal{E}(v)|}$ balances aggregation across all connected hyperedges, and
$\text{AGG}(\cdot)$ denotes an aggregation function that combines feature vectors from all nodes within each hyperedge into a unified representation \cite{feng2024hyper}.

Through this formulation, the hypergraph architecture naturally captures group-wise dependencies essential for industrial geometries, where structural precision depends on maintaining consistent relationships across multiple perspectives simultaneously, which represents a paradigm shift from local geometric optimization to global structural reasoning.

After hypergraph processing and mask-guided focus on object regions $\mathcal{M}$, our MVHG loss operates in the joint cross-view feature space:
\begin{equation}
\begin{aligned}
    \mathcal{L}_{\text{MVHG}} = \frac{1}{|\mathcal{M}|} \sum_{(h,w)\in\mathcal{M}} \Big\| &\bm{F}^{\text{masked}}_z[h,w,:] \\
    & - \bm{F}^{\text{masked}}_{\text{pred}}[h,w,:] \Big\|_2^2
\end{aligned}
\end{equation}
where $\bm{F}_z^{\text{masked}}$ and $\bm{F}_{\text{pred}}^{\text{masked}}$ represent the hypergraph-processed features from ground truth and predicted latents respectively, masked to focus on object regions. The step-by-step computation process is summarized in \Cref{alg:hypergraph_geometry_loss_vlined}.

\subsection{ForgeDreamer: Unified Industrial Text-to-3D Pipeline}
Having established semantic understanding through LoRA distillation and geometric modeling through hypergraph enhancement, we now integrate these components into a unified pipeline optimized for industrial applications. 

Our industrial-oriented pipeline employs a unified optimization objective that simultaneously addresses semantic understanding and geometric precision:
\begin{align}
    \mathcal{L}_{\text{total}} = \lambda_{\text{ISM}}\mathcal{L}_{\text{ISM}} + \lambda_{\text{MVHG}}\mathcal{L}_{\text{MVHG}}
\end{align}
where $\mathcal{L}_{\text{ISM}}$ ensures consistent pseudo-ground-truth generation through interval score matching, and $\mathcal{L}_{\text{MVHG}}$ enforces cross-view geometric consistency via our hypergraph formulation. This dual-objective design addresses the unique challenges of industrial 3D generation where both textual precision and geometric accuracy are critical.

Our framework utilizes 3D Gaussian Splatting for its superior efficiency in high-resolution rendering, essential for capturing intricate industrial details. The key insight is that our distilled LoRA weights improve the diffusion model's understanding of industrial product categories. The CVGCM module operates continuously throughout training, extracting higher-order geometric relationships across multiple viewpoints. We employ fixed multi-view sampling optimized for industrial objects, providing standardized geometric supervision that maintains manufacturing precision and dimensional accuracy across different component categories. The combination of distilled industrial knowledge and hypergraph geometric modeling ensures that these structural details are both visually coherent and geometrically consistent across all viewpoints.

\begin{figure*}[t]
\centering
\includegraphics[width=1.0\textwidth]{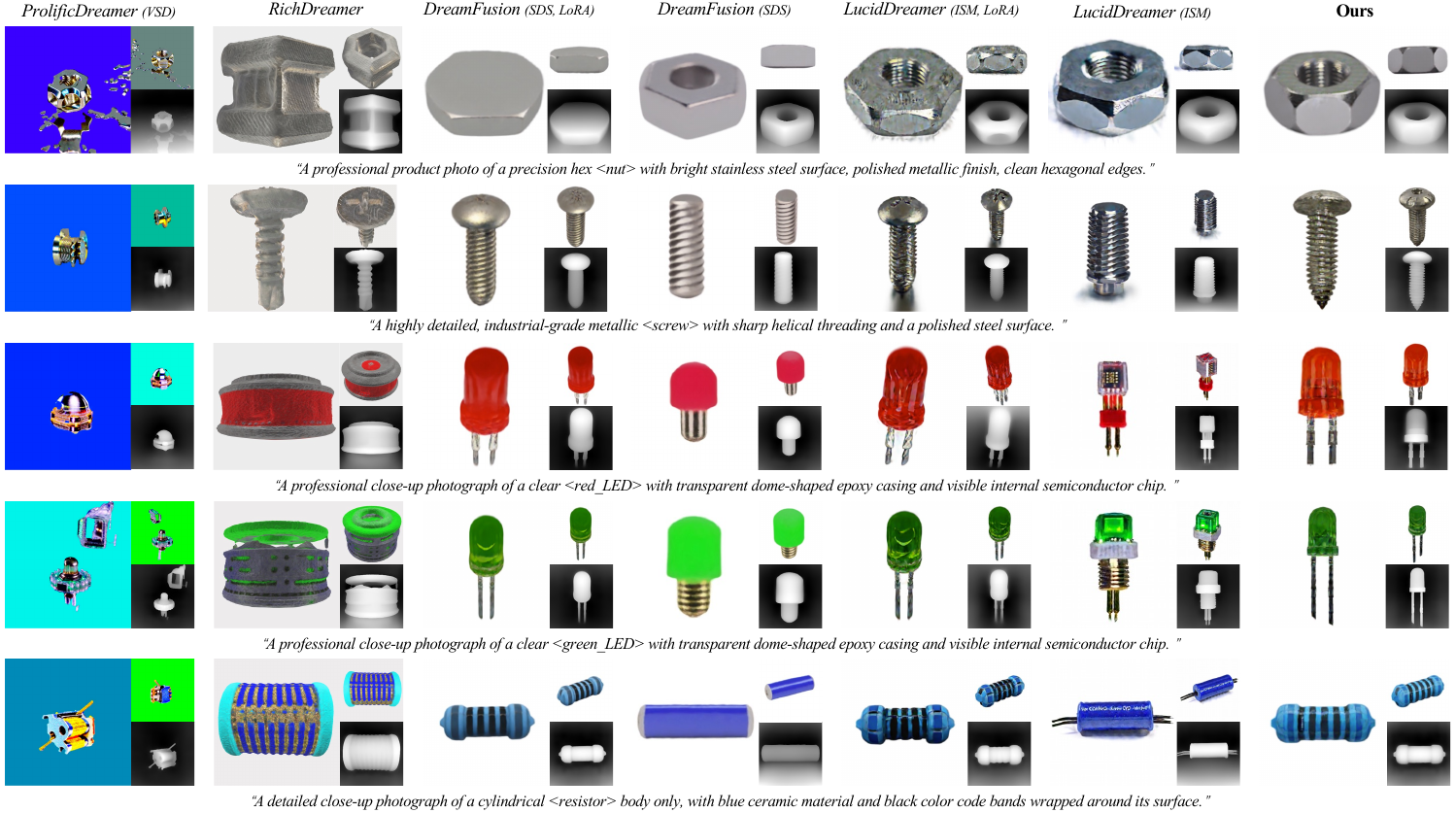} 
\caption{Qualitative Comparison with State-of-the-Art Methods. Visual results demonstrate the superior performance of our approach. See \textit{Appendix} for remaining categories.}
\label{fig4}
\end{figure*}

\begin{algorithm}[htb]
\KwIn{Image batch $\bm{X}$, prompt $c$, training steps $T$, batch size $B$}
\KwOut{$\mathcal{L}_{\text{MVHG}}$}
\BlankLine
\For{$t = 1$ \KwTo $T$}{
    Encode: $\bm{Z} = \text{VAE}(\bm{X})$\\
    Add noise: $\bm{Z}_{\text{noisy}} = \bm{Z} + \boldsymbol{\epsilon},\ \boldsymbol{\epsilon} \sim \mathcal{N}(0, \sigma^2)$\\
    Denoise: $\hat{\bm{X}}, \hat{\boldsymbol{\epsilon}} = \text{SD}(\bm{Z}_{\text{noisy}}, c)$\\
    Predict: $\bm{Z}_{\text{pred}} = \bm{Z}_{\text{noisy}} - \hat{\boldsymbol{\epsilon}}$\\
    Masks: $\bm{M}, \hat{\bm{M}} = \text{HSVMask}(\bm{X}),\ \text{HSVMask}(\hat{\bm{X}})$\\
    Concat: $\bm{z} = \text{concat}(\bm{Z}),\ \bm{z}_{\text{pred}} = \text{concat}(\bm{Z}_{\text{pred}})$\\
    Concat: $\bm{m} = \text{concat}(\bm{M}),\ \hat{\bm{m}} = \text{concat}(\hat{\bm{M}})$\\
    Hypergraphs: $H_{\bm{z}}, H_{\text{pred}} = \text{TopK}(\bm{z}),\ \text{TopK}(\bm{z}_{\text{pred}})$\\
    HGNN: $\bm{F}_{\bm{z}}, \bm{F}_{\text{pred}} = \text{HGNN}(H_{\bm{z}}),\ \text{HGNN}(H_{\text{pred}})$\\
    Masked feats: $\bm{F}_{\bm{z}}', \bm{F}_{\text{pred}}' = \bm{F}_{\bm{z}} \cdot \bm{m},\ \bm{F}_{\text{pred}} \cdot \hat{\bm{m}}$\\
    Loss: $\mathcal{L}_{\text{MVHG}} = \|\bm{F}_{\bm{z}}' - \bm{F}_{\text{pred}}'\|_2^2$
}
\BlankLine
\KwRet{$\mathcal{L}_{\mathrm{MVHG}}$}
\caption{$\mathcal{L}_{\text{MVHG}}$ Computation}
\label{alg:hypergraph_geometry_loss_vlined}
\end{algorithm}

In inference, our system operates through a coordinated multi-stage process that leverages both components synergistically. The pipeline begins with semantic-guided initialization, where text input is processed through our distilled LoRA-enhanced encoding to produce an initial 3D Gaussian representation. Subsequently, the system enters an iterative multi-view consistency enforcement loop where each iteration renders multi-view images from the current 3D Gaussian Splatting representation, applies CVGCM processing to capture cross-view dependencies, and updates the 3D parameters based on both semantic and geometric losses. 
\section{Experiments}
\label{sec:experiments}

\subsection{Dataset Construction and Specification}
To obtain reliable supervision for LoRA distillation, we constructed a small, controlled multi-view dataset covering ten industrial categories: six mechanical components (screw, nut, bearing, gasket, nail, and hexagonal stud) and four electronic components (ceramic capacitor, resistor, red LED, and green LED). Each category contains 20 high-resolution images captured from front and top perspectives. Further dataset details are provided in the \textit{Appendix}.

\subsection{Text-to-3D Generation}
To evaluate the overall effectiveness of our generation pipeline, we begin by visualizing representative 3D assets synthesized from textual prompts. As shown in \Cref{fig4}, our proposed method produces semantically accurate 3D shapes with rich structural fidelity. Compared to prior methods primarily designed for natural scene generation, our framework exhibits superior performance in terms of topological coherence, edge sharpness, and texture consistency. For instance, prompts such as “a metallic screw with clear threads” or “a red LED” yield geometries that not only align with human expectations but also preserve shape-related priors and text-driven attributes. This improvement is largely attributed to the integration of distilled LoRA weights and the proposed higher-order hypergraph geometric loss, both of which contribute to capturing fine-grained structural details and enhancing semantic alignment in the generation process.

\subsection{Qualitative Evaluation}
We evaluate generation quality from three perspectives: 
(1) \textbf{image-based qualitative comparison},  
(2) \textbf{quantitative evaluation} following the T3Bench~\cite{he2023t} protocol, and  
(3) \textbf{LLM-based judgment} using a large language model as an external evaluator.  

For qualitative comparison, we benchmark our method against several recent and representative Text-to-3D baselines, including ProlificDreamer~\cite{wang2023prolificdreamer} and RichDreamer~\cite{qiu2024richdreamer} in their original forms, as well as DreamFusion~\cite{poole2022dreamfusion} (with LoRA) and LucidDreamer~\cite{liang2024luciddreamer} (with LoRA), which are adapted with distilled LoRA weights equivalent to those used in our approach. This hybrid comparison setup allows us to assess both the baseline performance of existing models and the impact of integrating our LoRA distillation strategy under consistent conditions. A condensed quantitative comparison of processing time and average quality scores is provided in \Cref{tab:condensed_avg_sorted_aligned}. All experiments are conducted on an NVIDIA RTX 4090 to ensure computational fairness. Representative qualitative results are shown in \Cref{fig4}, with additional comparisons on natural scenes provided in the \textit{Appendix}.

\begin{table}[h!]
\centering
\caption{Condensed comparison of average scores (10 categories). $^{\dagger}$Total processing time; $^{*}$Average T3Bench~\cite{he2023t} quality score. Best and second-best scores are \textbf{bolded} and \underline{underlined}, respectively. See \textit{Appendix} for full results.}
\label{tab:condensed_avg_sorted_aligned}
\resizebox{\columnwidth}{!}{%
\begin{tabular}{llc}
\toprule
\textbf{Method} & \textbf{Time}$^{\dagger}$ & \textbf{Average}$^{*}$ \\
\midrule
ProlificDreamer \cite{wang2023prolificdreamer} (w/o LoRA) & $\sim$10 hours & 25.13 \\
DreamFusion \cite{poole2022dreamfusion} (w/o LoRA)      & \multirow{2}{*}{6 hours} & 41.91 \\
DreamFusion \cite{poole2022dreamfusion} (w/ LoRA) & & 44.83 \\
RichDreamer \cite{qiu2024richdreamer} (w/o LoRA)       & 120 minutes & 28.27 \\
LucidDreamer \cite{liang2024luciddreamer} (w/o LoRA)      & \multirow{2}{*}{110 minutes} & \underline{47.10} \\
LucidDreamer \cite{liang2024luciddreamer} (w/ LoRA) & & 46.75 \\
\midrule
Ours                              & 130 minutes & \textbf{50.88} \\
\bottomrule
\end{tabular}%
}
\end{table}


\begin{figure}[h!]
\centering
\includegraphics[width=0.9\columnwidth]{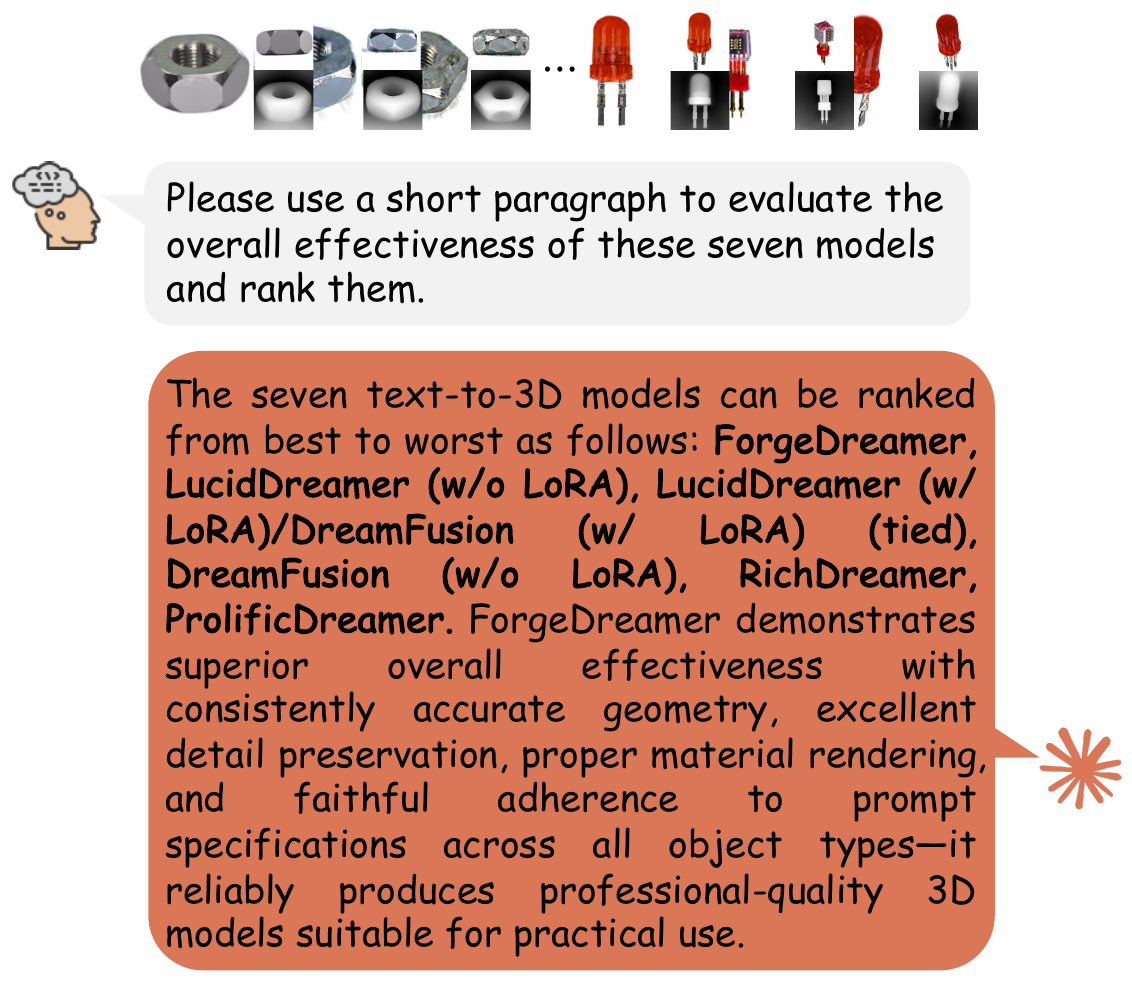} 
\caption{
LLM-based qualitative ranking across ten object prompts, showing that our method achieves the highest overall fidelity and consistency. 
Full evaluation results are included in the \textit{Appendix}.
}
\label{fig5}
\end{figure}

Across all ten categories, our method consistently produces exceptionally high-fidelity geometry, intricate detailed structures, photo-realistic materials, and outstanding prompt adherence, resulting in highly reliable and practical 3D asset generation suitable for downstream use. In notoriously demanding industrial-specific categories such as bolts, nuts, and connectors, our model notably and robustly preserves accurate topology and successfully avoids the typical artifacts such as disconnected geometry or unnatural surface oversmoothing that plague other methods. The rigorous LLM-based comparative evaluation, which statistically confirms these advantages, is summarized in \Cref{fig5}.



\subsection{Fine-tuning Strategy Analysis}
To evaluate different fine-tuning paradigms, we compare three variants of our framework. \textbf{No fine-tuning:} The base model without any domain-specific adaptation. \textbf{Single-LoRA adaptation:} Per-concept LoRA modules trained independently. \textbf{Multi-LoRA distillation:} Our proposed strategy merges multiple LoRAs into a unified model.

\begin{figure}[h!]
\centering
\includegraphics[width=0.88\columnwidth]{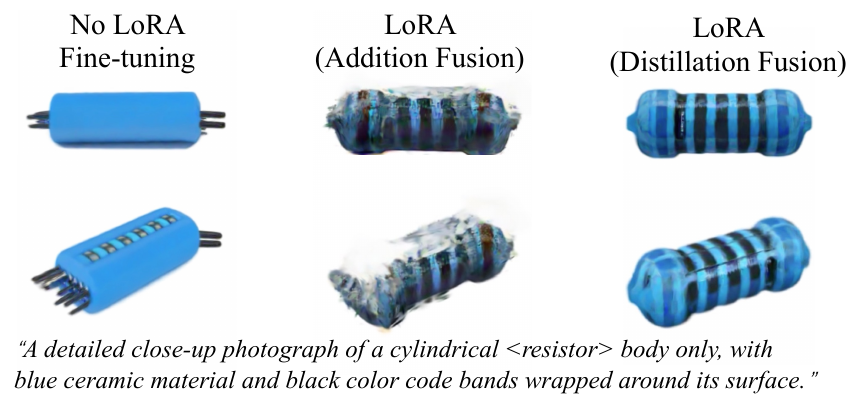} 
\caption{Multi-Expert LoRA Integration: Addition Fusion vs. Distillation Fusion}
\label{fig6}
\vspace{-5mm}
\end{figure}

\Cref{fig6} shows qualitative differences: the base model lacks fine structural detail, while Single-LoRA improves fidelity but struggles with generalization. Our Multi-LoRA distillation achieves the best trade-off, producing sharper geometry and consistent semantics across complex prompts.
\Cref{tab:concept_preservation_compressed} complements this with quantitative evidence, showing that distillation maintains higher cosine similarity to original LoRAs compared to simple addition, especially as the number of fused LoRAs increases.

\begin{table}[h!]
\centering
\caption{Comparison of Average Concept Preservation Scores evaluated by CLIP-ViT-L/14. The full detailed table is presented in the \textit{Appendix}.}
\label{tab:concept_preservation_compressed}

\resizebox{\columnwidth}{!}{
    \begin{tabular}{lccc}
    \toprule
    \textbf{Method} & \textbf{Two LoRAs} & \textbf{Four LoRAs} & \textbf{Six LoRAs} \\
    \midrule
    Addition      & 0.938 & 0.814 & 0.633 \\
    Distillation  & \textbf{0.965} & \textbf{0.949} & \textbf{0.952} \\
    \bottomrule
    \end{tabular}
}
\end{table}

\subsection{Ablation Study}
\subsubsection{Effectiveness of MVHG Loss}
\begin{figure}[t!]
\centering
\includegraphics[width=0.85\columnwidth]{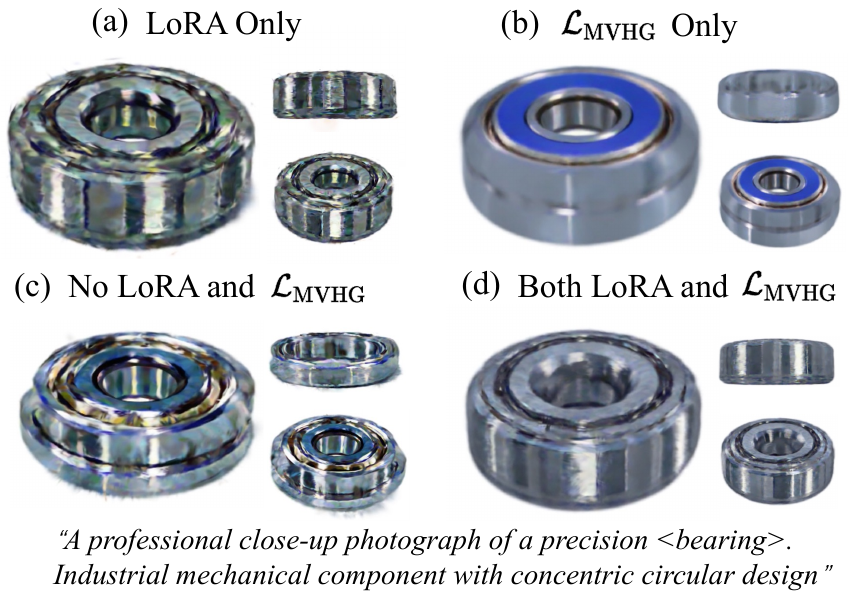} 
\caption{Component-wise Ablation Analysis: LoRA (Distillation Fusion) vs. MVHG Loss vs. Combined Approach}
\label{fig8}
\vspace{-5mm}
\end{figure}

 As shown in \Cref{fig8}, incorporating MVHG loss significantly improves the geometric fidelity and spatial consistency of the generated 3D shapes. Without it, models tend to produce artifacts such as inconsistent topology across views and distortions in fine structures. The MVHG loss introduces higher-order geometric constraints that capture inter-view correlations via a hypergraph representation, effectively regularizing complex surfaces and preserving overall shape integrity. These improvements underscore the importance of MVHG loss in enhancing structural coherence and reducing inter-view discrepancies in the synthesized results.

\subsubsection{Influence of Distilled LoRA Fine-tuning}
We further assess the impact of our proposed Distilled LoRA fine-tuning approach on the overall performance of the text-to-3D generation pipeline. As illustrated in \Cref{fig8}, the absence of Distilled LoRA leads to a marked drop in visual plausibility and semantic alignment with input text prompts. Notably, models without our distillation framework often struggle to capture fine-grained attributes in underrepresented categories and exhibit unstable geometry when dealing with complex or multi-part semantics. In contrast, our Distilled LoRA effectively consolidates knowledge from multiple expert LoRAs, enhancing the model’s ability to encode domain-specific priors and adapt to subtle textual variations. This leads to more faithful and structurally coherent outputs. 
\section{Conclusion}
\label{sec:conclusion}

In this paper, we present a novel framework, ForgeDreamer, for high-fidelity industrial text-to-3D generation that addresses two critical limitations of existing methods: domain adaptation challenges and geometric reasoning deficiencies. Our approach introduces two key innovations through a systematic progression: a Multi-Expert LoRA Geometric mechanism and a Cross-View Hypergraph Semantic Enhancement. The key contribution is to improve geometric precision through multi-expert LoRA, enabling more effective semantic modeling, while cross-view hypergraph understanding ensures precise geometric relationships across multiple views. Extensive experiments on our custom industrial dataset demonstrate superior performance compared to state-of-the-art methods, achieving enhanced semantic generalization and geometric fidelity for industrial components in various practical engineering applications.
\section*{Acknowledgements}

This research work was financially supported in part by the Guangdong Major Project of Basic Research under Grant 2023B0303000009, in part by the Shenzhen Fundamental Research Fund under Grant JCYJ20230808105212023, in part by the NSFC Youth Fund Project under Grant 62403326, in part by the Research Team Cultivation Program of ShenZhen University under Grant 2023JCT004, and in part by the Shenzhen University 2035 Program for Excellent Research under Grant 00000224.
{
    \small
    \bibliographystyle{ieeenat_fullname}
    \bibliography{main}

@String(TOG= {ACM Trans. Graph.})

@String(ICASSP=	{ICASSP})

@String(ICLR = {Int. Conf. Learn. Represent.})

@String(AAAI = {AAAI})

@String(TOG   = {ACM TOG})

@String(ICLR  = {ICLR})

@article{gao2022hgnn+,
  title={Hgnn+: General hypergraph neural networks},
  author={Gao, Yue and Feng, Yifan and Ji, Shuyi and Ji, Rongrong},
  journal={IEEE Transactions on Pattern Analysis and Machine Intelligence},
  volume={45},
  number={3},
  pages={3181--3199},
  year={2022},
  publisher={IEEE}
}

@article{feng2024hyper,
  title={Hyper-yolo: When visual object detection meets hypergraph computation},
  author={Feng, Yifan and Huang, Jiangang and Du, Shaoyi and Ying, Shihui and Yong, Jun-Hai and Li, Yipeng and Ding, Guiguang and Ji, Rongrong and Gao, Yue},
  journal={IEEE Transactions on Pattern Analysis and Machine Intelligence},
  year={2024},
  publisher={IEEE}
}

@inproceedings{genova2020local,
  title={Local deep implicit functions for 3d shape},
  author={Genova, Kyle and Cole, Forrester and Sud, Avneesh and Sarna, Aaron and Funkhouser, Thomas},
  booktitle={Proceedings of the IEEE/CVF conference on computer vision and pattern recognition},
  pages={4857--4866},
  year={2020}
}

@inproceedings{groueix2018papier,
  title={A papier-m{\^a}ch{\'e} approach to learning 3d surface generation},
  author={Groueix, Thibault and Fisher, Matthew and Kim, Vladimir G and Russell, Bryan C and Aubry, Mathieu},
  booktitle={Proceedings of the IEEE conference on computer vision and pattern recognition},
  pages={216--224},
  year={2018}
}

@article{schwarz2020graf,
  title={Graf: Generative radiance fields for 3d-aware image synthesis},
  author={Schwarz, Katja and Liao, Yiyi and Niemeyer, Michael and Geiger, Andreas},
  journal={Advances in neural information processing systems},
  volume={33},
  pages={20154--20166},
  year={2020}
}

@inproceedings{niemeyer2021giraffe,
  title={Giraffe: Representing scenes as compositional generative neural feature fields},
  author={Niemeyer, Michael and Geiger, Andreas},
  booktitle={Proceedings of the IEEE/CVF conference on computer vision and pattern recognition},
  pages={11453--11464},
  year={2021}
}

@inproceedings{wang2021ibrnet,
  title={Ibrnet: Learning multi-view image-based rendering},
  author={Wang, Qianqian and Wang, Zhicheng and Genova, Kyle and Srinivasan, Pratul P and Zhou, Howard and Barron, Jonathan T and Martin-Brualla, Ricardo and Snavely, Noah and Funkhouser, Thomas},
  booktitle={Proceedings of the IEEE/CVF conference on computer vision and pattern recognition},
  pages={4690--4699},
  year={2021}
}

@inproceedings{chen2021mvsnerf,
  title={Mvsnerf: Fast generalizable radiance field reconstruction from multi-view stereo},
  author={Chen, Anpei and Xu, Zexiang and Zhao, Fuqiang and Zhang, Xiaoshuai and Xiang, Fanbo and Yu, Jingyi and Su, Hao},
  booktitle={Proceedings of the IEEE/CVF international conference on computer vision},
  pages={14124--14133},
  year={2021}
}

@inproceedings{yu2021pixelnerf,
  title={pixelnerf: Neural radiance fields from one or few images},
  author={Yu, Alex and Ye, Vickie and Tancik, Matthew and Kanazawa, Angjoo},
  booktitle={Proceedings of the IEEE/CVF conference on computer vision and pattern recognition},
  pages={4578--4587},
  year={2021}
}

@inproceedings{kumari2023multi,
  title={Multi-concept customization of text-to-image diffusion},
  author={Kumari, Nupur and Zhang, Bingliang and Zhang, Richard and Shechtman, Eli and Zhu, Jun-Yan},
  booktitle={Proceedings of the IEEE/CVF conference on computer vision and pattern recognition},
  pages={1931--1941},
  year={2023}
}

@article{zhang2023adalora,
  title={Adalora: Adaptive budget allocation for parameter-efficient fine-tuning},
  author={Zhang, Qingru and Chen, Minshuo and Bukharin, Alexander and Karampatziakis, Nikos and He, Pengcheng and Cheng, Yu and Chen, Weizhu and Zhao, Tuo},
  journal={arXiv preprint arXiv:2303.10512},
  year={2023}
}

@article{dettmers2023qlora,
  title={Qlora: Efficient finetuning of quantized llms, 2023},
  author={Dettmers, Tim and Pagnoni, Artidoro and Holtzman, Ari and Zettlemoyer, Luke},
  journal={URL https://arxiv. org/abs/2305.14314},
  volume={2},
  year={2023}
}

@inproceedings{lambourne2021brepnet,
  title={Brepnet: A topological message passing system for solid models},
  author={Lambourne, Joseph G and Willis, Karl DD and Jayaraman, Pradeep Kumar and Sanghi, Aditya and Meltzer, Peter and Shayani, Hooman},
  booktitle={Proceedings of the IEEE/CVF conference on computer vision and pattern recognition},
  pages={12773--12782},
  year={2021}
}

@inproceedings{willis2022joinable,
  title={Joinable: Learning bottom-up assembly of parametric cad joints},
  author={Willis, Karl DD and Jayaraman, Pradeep Kumar and Chu, Hang and Tian, Yunsheng and Li, Yifei and Grandi, Daniele and Sanghi, Aditya and Tran, Linh and Lambourne, Joseph G and Solar-Lezama, Armando and others},
  booktitle={Proceedings of the IEEE/CVF conference on computer vision and pattern recognition},
  pages={15849--15860},
  year={2022}
}

@inproceedings{zhou20213d,
  title={3d shape generation and completion through point-voxel diffusion},
  author={Zhou, Linqi and Du, Yilun and Wu, Jiajun},
  booktitle={Proceedings of the IEEE/CVF international conference on computer vision},
  pages={5826--5835},
  year={2021}
}

@inproceedings{luo2021diffusion,
  title={Diffusion probabilistic models for 3d point cloud generation},
  author={Luo, Shitong and Hu, Wei},
  booktitle={Proceedings of the IEEE/CVF conference on computer vision and pattern recognition},
  pages={2837--2845},
  year={2021}
}

@article{hong2023lrm,
  title={Lrm: Large reconstruction model for single image to 3d},
  author={Hong, Yicong and Zhang, Kai and Gu, Jiuxiang and Bi, Sai and Zhou, Yang and Liu, Difan and Liu, Feng and Sunkavalli, Kalyan and Bui, Trung and Tan, Hao},
  journal={arXiv preprint arXiv:2311.04400},
  year={2023}
}

@article{shi2023mvdream,
  title={Mvdream: Multi-view diffusion for 3d generation},
  author={Shi, Yichun and Wang, Peng and Ye, Jianglong and Long, Mai and Li, Kejie and Yang, Xiao},
  journal={arXiv preprint arXiv:2308.16512},
  year={2023}
}

@inproceedings{liu2023zero,
  title={Zero-1-to-3: Zero-shot one image to 3d object},
  author={Liu, Ruoshi and Wu, Rundi and Van Hoorick, Basile and Tokmakov, Pavel and Zakharov, Sergey and Vondrick, Carl},
  booktitle={Proceedings of the IEEE/CVF international conference on computer vision},
  pages={9298--9309},
  year={2023}
}

@article{muller2022instant,
  title={Instant neural graphics primitives with a multiresolution hash encoding},
  author={M{\"u}ller, Thomas and Evans, Alex and Schied, Christoph and Keller, Alexander},
  journal={ACM transactions on graphics (TOG)},
  volume={41},
  number={4},
  pages={1--15},
  year={2022},
  publisher={ACM New York, NY, USA}
}

@article{ramesh2022hierarchical,
  title={Hierarchical text-conditional image generation with clip latents},
  author={Ramesh, Aditya and Dhariwal, Prafulla and Nichol, Alex and Chu, Casey and Chen, Mark},
  journal={arXiv preprint arXiv:2204.06125},
  volume={1},
  number={2},
  pages={3},
  year={2022}
}

@inproceedings{metzer2023latent,
  title={Latent-nerf for shape-guided generation of 3d shapes and textures},
  author={Metzer, Gal and Richardson, Elad and Patashnik, Or and Giryes, Raja and Cohen-Or, Daniel},
  booktitle={Proceedings of the IEEE/CVF conference on computer vision and pattern recognition},
  pages={12663--12673},
  year={2023}
}

@inproceedings{chen2023fantasia3d,
  title={Fantasia3d: Disentangling geometry and appearance for high-quality text-to-3d content creation},
  author={Chen, Rui and Chen, Yongwei and Jiao, Ningxin and Jia, Kui},
  booktitle={Proceedings of the IEEE/CVF international conference on computer vision},
  pages={22246--22256},
  year={2023}
}

@inproceedings{lin2023magic3d,
  title={Magic3d: High-resolution text-to-3d content creation},
  author={Lin, Chen-Hsuan and Gao, Jun and Tang, Luming and Takikawa, Towaki and Zeng, Xiaohui and Huang, Xun and Kreis, Karsten and Fidler, Sanja and Liu, Ming-Yu and Lin, Tsung-Yi},
  booktitle={Proceedings of the IEEE/CVF conference on computer vision and pattern recognition},
  pages={300--309},
  year={2023}
}

@article{wang2021nerf,
  title={NeRF--: Neural radiance fields without known camera parameters},
  author={Wang, Zirui and Wu, Shangzhe and Xie, Weidi and Chen, Min and Prisacariu, Victor Adrian},
  year={2021}
}

@article{wu2024recent,
  title={Recent advances in 3d gaussian splatting},
  author={Wu, Tong and Yuan, Yu-Jie and Zhang, Ling-Xiao and Yang, Jie and Cao, Yan-Pei and Yan, Ling-Qi and Gao, Lin},
  journal={Computational Visual Media},
  volume={10},
  number={4},
  pages={613--642},
  year={2024},
  publisher={TUP}
}

@article{mildenhall2021nerf,
  title={Nerf: Representing scenes as neural radiance fields for view synthesis},
  author={Mildenhall, Ben and Srinivasan, Pratul P and Tancik, Matthew and Barron, Jonathan T and Ramamoorthi, Ravi and Ng, Ren},
  journal={Communications of the ACM},
  volume={65},
  number={1},
  pages={99--106},
  year={2021},
  publisher={ACM New York, NY, USA}
}

@article{poole2022dreamfusion,
  title={Dreamfusion: Text-to-3d using 2d diffusion},
  author={Poole, Ben and Jain, Ajay and Barron, Jonathan T and Mildenhall, Ben},
  journal={arXiv preprint arXiv:2209.14988},
  year={2022}
}

@inproceedings{liang2024luciddreamer,
  title={Luciddreamer: Towards high-fidelity text-to-3d generation via interval score matching},
  author={Liang, Yixun and Yang, Xin and Lin, Jiantao and Li, Haodong and Xu, Xiaogang and Chen, Yingcong},
  booktitle={Proceedings of the IEEE/CVF conference on computer vision and pattern recognition},
  pages={6517--6526},
  year={2024}
}

@article{wang2023prolificdreamer,
  title={Prolificdreamer: High-fidelity and diverse text-to-3d generation with variational score distillation},
  author={Wang, Zhengyi and Lu, Cheng and Wang, Yikai and Bao, Fan and Li, Chongxuan and Su, Hang and Zhu, Jun},
  journal={Advances in neural information processing systems},
  volume={36},
  pages={8406--8441},
  year={2023}
}

@article{hu2022lora,
  title={Lora: Low-rank adaptation of large language models.},
  author={Hu, Edward J and Shen, Yelong and Wallis, Phillip and Allen-Zhu, Zeyuan and Li, Yuanzhi and Wang, Shean and Wang, Lu and Chen, Weizhu and others},
  journal={ICLR},
  volume={1},
  number={2},
  pages={3},
  year={2022}
}

@inproceedings{rombach2022high,
  title={High-resolution image synthesis with latent diffusion models},
  author={Rombach, Robin and Blattmann, Andreas and Lorenz, Dominik and Esser, Patrick and Ommer, Bj{\"o}rn},
  booktitle={Proceedings of the IEEE/CVF conference on computer vision and pattern recognition},
  pages={10684--10695},
  year={2022}
}

@article{ho2020denoising,
  title={Denoising diffusion probabilistic models},
  author={Ho, Jonathan and Jain, Ajay and Abbeel, Pieter},
  journal={Advances in neural information processing systems},
  volume={33},
  pages={6840--6851},
  year={2020}
}

@article{zhong2024multi,
  title={Multi-lora composition for image generation},
  author={Zhong, Ming and Shen, Yelong and Wang, Shuohang and Lu, Yadong and Jiao, Yizhu and Ouyang, Siru and Yu, Donghan and Han, Jiawei and Chen, Weizhu},
  journal={arXiv preprint arXiv:2402.16843},
  year={2024}
}

@inproceedings{feng2019hypergraph,
  title={Hypergraph neural networks},
  author={Feng, Yifan and You, Haoxuan and Zhang, Zizhao and Ji, Rongrong and Gao, Yue},
  booktitle={Proceedings of the AAAI conference on artificial intelligence},
  volume={33},
  number={01},
  pages={3558--3565},
  year={2019}
}

@article{di2025hyper,
  title={Hyper-3dg: Text-to-3d gaussian generation via hypergraph},
  author={Di, Donglin and Yang, Jiahui and Luo, Chaofan and Xue, Zhou and Chen, Wei and Yang, Xun and Gao, Yue},
  journal={International Journal of Computer Vision},
  volume={133},
  number={5},
  pages={2886--2909},
  year={2025},
  publisher={Springer}
}

@article{kerbl20233d,
  title={3D Gaussian splatting for real-time radiance field rendering.},
  author={Kerbl, Bernhard and Kopanas, Georgios and Leimk{\"u}hler, Thomas and Drettakis, George},
  journal={ACM Trans. Graph.},
  volume={42},
  number={4},
  pages={139--1},
  year={2023}
}

@article{li2024era3d,
  title={Era3d: High-resolution multiview diffusion using efficient row-wise attention},
  author={Li, Peng and Liu, Yuan and Long, Xiaoxiao and Zhang, Feihu and Lin, Cheng and Li, Mengfei and Qi, Xingqun and Zhang, Shanghang and Xue, Wei and Luo, Wenhan and others},
  journal={Advances in Neural Information Processing Systems},
  volume={37},
  pages={55975--56000},
  year={2024}
}

@inproceedings{qiu2024richdreamer,
  title={Richdreamer: A generalizable normal-depth diffusion model for detail richness in text-to-3d},
  author={Qiu, Lingteng and Chen, Guanying and Gu, Xiaodong and Zuo, Qi and Xu, Mutian and Wu, Yushuang and Yuan, Weihao and Dong, Zilong and Bo, Liefeng and Han, Xiaoguang},
  booktitle={Proceedings of the IEEE/CVF conference on computer vision and pattern recognition},
  pages={9914--9925},
  year={2024}
}

@article{nichol2022point,
  title={Point-e: A system for generating 3d point clouds from complex prompts},
  author={Nichol, Alex and Jun, Heewoo and Dhariwal, Prafulla and Mishkin, Pamela and Chen, Mark},
  journal={arXiv preprint arXiv:2212.08751},
  year={2022}
}

@inproceedings{liu2024pi3d,
  title={Pi3d: Efficient text-to-3d generation with pseudo-image diffusion},
  author={Liu, Ying-Tian and Guo, Yuan-Chen and Luo, Guan and Sun, Heyi and Yin, Wei and Zhang, Song-Hai},
  booktitle={Proceedings of the IEEE/CVF Conference on Computer Vision and Pattern Recognition},
  pages={19915--19924},
  year={2024}
}

@inproceedings{zhou2024dreampropeller,
  title={Dreampropeller: Supercharge text-to-3d generation with parallel sampling},
  author={Zhou, Linqi and Shih, Andy and Meng, Chenlin and Ermon, Stefano},
  booktitle={Proceedings of the IEEE/CVF Conference on Computer Vision and Pattern Recognition},
  pages={4610--4619},
  year={2024}
}

@inproceedings{wu2024consistent3d,
  title={Consistent3d: Towards consistent high-fidelity text-to-3d generation with deterministic sampling prior},
  author={Wu, Zike and Zhou, Pan and Yi, Xuanyu and Yuan, Xiaoding and Zhang, Hanwang},
  booktitle={Proceedings of the IEEE/CVF Conference on Computer Vision and Pattern Recognition},
  pages={9892--9902},
  year={2024}
}

@inproceedings{raj2023dreambooth3d,
  title={Dreambooth3d: Subject-driven text-to-3d generation},
  author={Raj, Amit and Kaza, Srinivas and Poole, Ben and Niemeyer, Michael and Ruiz, Nataniel and Mildenhall, Ben and Zada, Shiran and Aberman, Kfir and Rubinstein, Michael and Barron, Jonathan and others},
  booktitle={Proceedings of the IEEE/CVF international conference on computer vision},
  pages={2349--2359},
  year={2023}
}

@article{macreating,
  title={Creating High-quality 3D Content by Bridging the Gap Between Text-to-2D and Text-to-3D Generation},
  author={Ma, Yiwei and Fan, Yijun and Ji, Jiayi and Wang, Haowei and Yin, Haibing and Sun, Xiaoshuai and Ji, Rongrong},
  journal={ACM Transactions on Multimedia Computing, Communications and Applications},
  publisher={ACM New York, NY},
  year={2024}
}

@inproceedings{huang2024dreamcontrol,
  title={Dreamcontrol: Control-based text-to-3d generation with 3d self-prior},
  author={Huang, Tianyu and Zeng, Yihan and Zhang, Zhilu and Xu, Wan and Xu, Hang and Xu, Songcen and Lau, Rynson WH and Zuo, Wangmeng},
  booktitle={Proceedings of the IEEE/CVF conference on computer vision and pattern recognition},
  pages={5364--5373},
  year={2024}
}

@article{he2023t,
  title={T\textsuperscript{3} Bench: Benchmarking Current Progress in Text-to-3D Generation},
  author={He, Yuze and Bai, Yushi and Lin, Matthieu and Zhao, Wang and Hu, Yubin and Sheng, Jenny and Yi, Ran and Li, Juanzi and Liu, Yong-Jin},
  journal={arXiv preprint arXiv:2310.02977},
  year={2023}
}

@inproceedings{wang2024real,
  title={Real-iad: A real-world multi-view dataset for benchmarking versatile industrial anomaly detection},
  author={Wang, Chengjie and Zhu, Wenbing and Gao, Bin-Bin and Gan, Zhenye and Zhang, Jiangning and Gu, Zhihao and Qian, Shuguang and Chen, Mingang and Ma, Lizhuang},
  booktitle={Proceedings of the IEEE/CVF Conference on Computer Vision and Pattern Recognition},
  pages={22883--22892},
  year={2024}
}

@article{bergmann2021mvtec,
  title={The mvtec 3d-ad dataset for unsupervised 3d anomaly detection and localization},
  author={Bergmann, Paul and Jin, Xin and Sattlegger, David and Steger, Carsten},
  journal={arXiv preprint arXiv:2112.09045},
  year={2021}
}

@article{zuo2025padiff,
  title={PADiff: Reconstruction From Patch to Pixel With Normality-Guided Diffusion Model for Unsupervised Anomaly Localization},
  author={Zuo, Zuo and Dong, Jiahao and Wu, Yao and Qu, Yanyun and Wu, Zongze},
  journal={IEEE Transactions on Neural Networks and Learning Systems},
  year={2025},
  publisher={IEEE}
}

@inproceedings{zuo2024reconstruction,
  title={A reconstruction-based feature adaptation for anomaly detection with self-supervised multi-scale aggregation},
  author={Zuo, Zuo and Wu, Zongze and Chen, Badong and Zhong, Xiaopin},
  booktitle={ICASSP 2024-2024 IEEE International Conference on Acoustics, Speech and Signal Processing (ICASSP)},
  pages={5840--5844},
  year={2024},
  organization={IEEE}
}

@article{luo2022lattice,
  title={Lattice network for lightweight image restoration},
  author={Luo, Xiaotong and Qu, Yanyun and Xie, Yuan and Zhang, Yulun and Li, Cuihua and Fu, Yun},
  journal={IEEE Transactions on Pattern Analysis and Machine Intelligence},
  volume={45},
  number={4},
  pages={4826--4842},
  year={2022},
  publisher={IEEE}
}

@inproceedings{luo2024skipdiff,
  title={Skipdiff: Adaptive skip diffusion model for high-fidelity perceptual image super-resolution},
  author={Luo, Xiaotong and Xie, Yuan and Qu, Yanyun and Fu, Yun},
  booktitle={Proceedings of the AAAI Conference on Artificial Intelligence},
  volume={38},
  number={5},
  pages={4017--4025},
  year={2024}
}
}
 
\clearpage
\setcounter{page}{1}
\maketitlesupplementary


\section{Appendix}

\subsection{Code and Dataset Examples}
We provide example implementations of the key components of our method, ForgeDreamer—including the LoRA distillation pipeline and the cross-view hypergraph enhancement module—in the supplementary materials to facilitate reproducibility. In addition, sample data from our multi-view industrial dataset are also included in the supplement to illustrate the data format and preprocessing workflow.

\subsection{Limitations of Existing Public 3D Datasets}
Although widely used in industrial anomaly detection and 3D perception, existing public datasets such as MVTec 3D-AD and Real-IAD are not well suited for our LoRA distillation framework. MVTec 3D-AD provides RGB images paired with depth or point cloud data; however, these point clouds are often incomplete. When generating a front-view image, the lower portion of the geometry is frequently missing, making it impossible to obtain a consistent and fully visible front-view observation required for LoRA training.

Real-IAD suffers from a different limitation: the dataset supplies only two viewpoints for each object—an oblique \(45^{\circ}\) view and a strictly top-down view. Since no true front-view images are available, the dataset cannot support the dual-perspective (front and up) supervision that our method relies on to ensure stable and geometry-consistent LoRA adaptation.

\begin{figure}[t!]
\centering
\includegraphics[width=1.0\columnwidth]{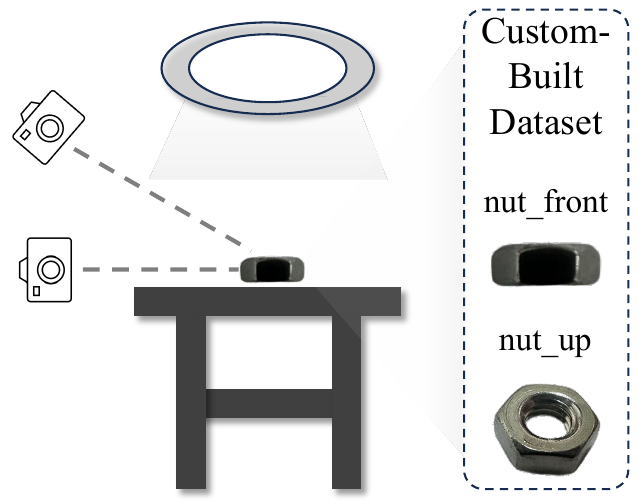} 
\caption{Our data capture apparatus (left) and collected dataset samples (right). The setup utilizes controlled lighting and fixed mounts to acquire consistent front and top-down view images for each industrial component.}
\label{fig:1}
\end{figure}

\begin{figure*}[t!]
    \centering

    \begin{subfigure}[b]{0.95\textwidth}
        \centering
        \includegraphics[width=\textwidth]{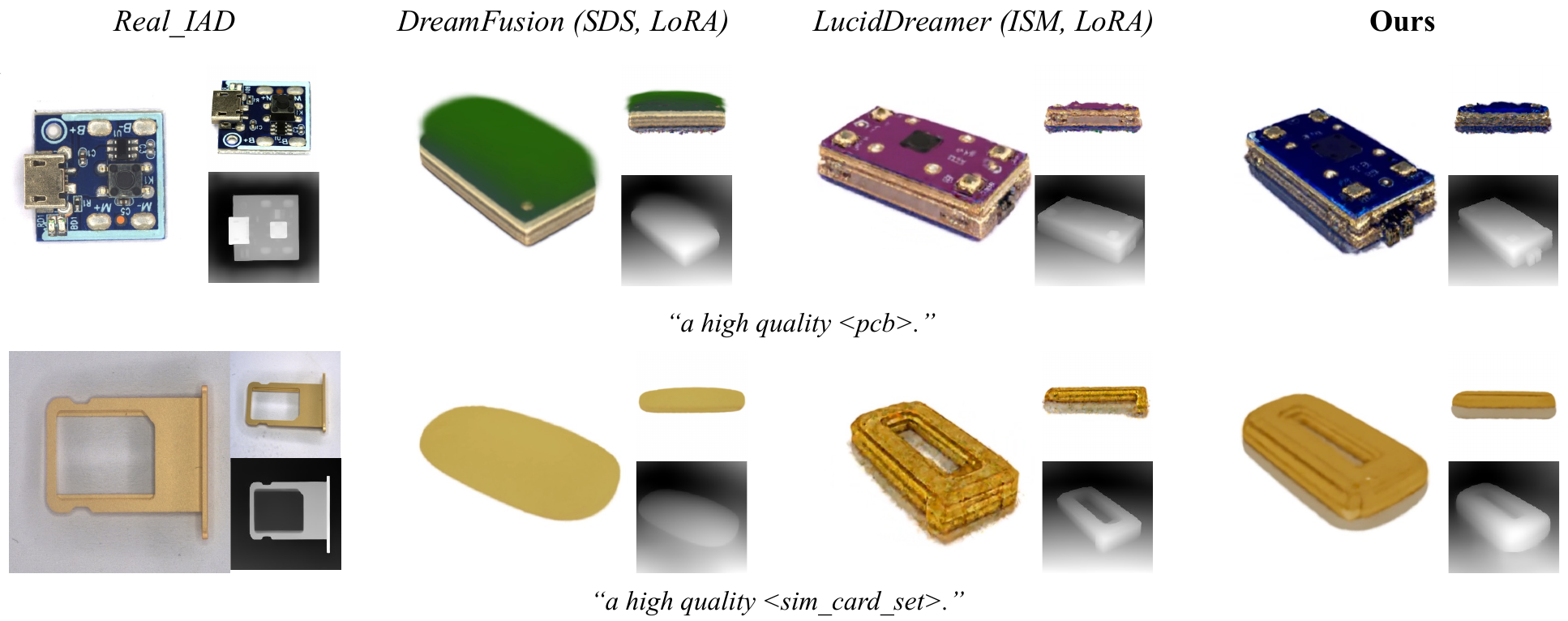}
        \caption{Results on the Real-IAD dataset. Our method successfully identifies and localizes industrial defects.}
        \label{fig:dataset_real_iad}
    \end{subfigure}

    \vspace{8pt}

    \begin{subfigure}[b]{0.95\textwidth}
        \centering
        \includegraphics[width=\textwidth]{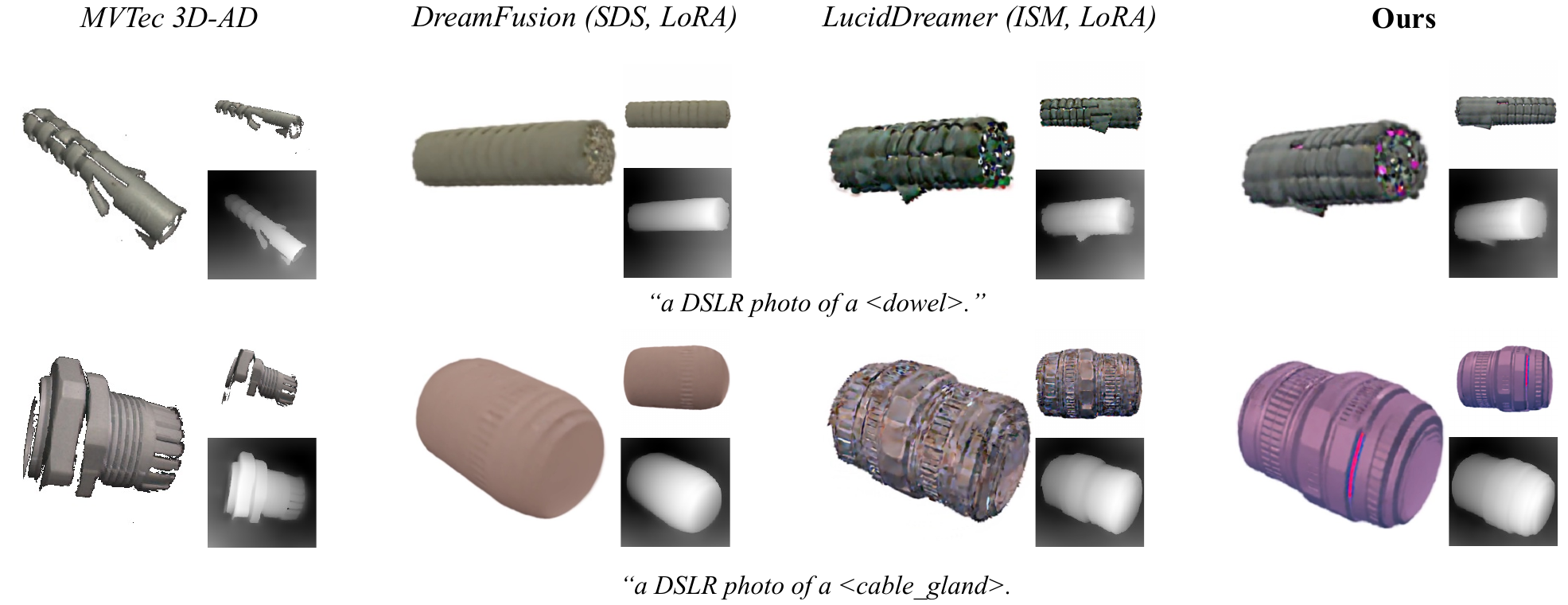}
        \caption{Comparison on the MVTec 3D-AD dataset. Our approach provides superior defect coverage compared to prior methods.}
        \label{fig:dataset_mvtec_3d}
    \end{subfigure}

    \caption{
        Qualitative results on two public industrial datasets. 
        (a) Real-IAD: our method accurately detects and localizes defects. 
        (b) MVTec 3D-AD: our framework achieves improved defect completeness and surface consistency.
    }
    \label{fig:combined_public_datasets}
\end{figure*}

To validate these limitations empirically, we conducted experiments using both datasets. As shown in \Cref{fig:dataset_real_iad} and \Cref{fig:dataset_mvtec_3d}, models trained on MVTec 3D-AD and Real-IAD exhibit degraded 3D generation quality, including incomplete geometry reconstruction, unstable texture synthesis, and inconsistent cross-view appearance. These results further demonstrate that the characteristics of existing datasets prevent them from providing the clean, multi-view supervision required by our framework, thereby motivating the construction of our own dataset.

\subsection{Dataset Construction and Specification}
To train and evaluate our LoRA distillation framework, we constructed a multi-view dataset comprising ten object categories: six mechanical components (screw, nut, bearing, gasket, nail, hexagonal stud) and four electronic components (ceramic capacitor, resistor, red LED, green LED). The dataset contains more than 200 high-resolution images (512×512 pixels), with 20 images per category evenly captured from front and up view perspectives. The data were collected using a standardized imaging setup equipped with a uniform ring-light illumination system, as illustrated schematically in \Cref{fig:1}.

For each object category, we captured 10 front view images with objects positioned to show their primary functional surface, and 10 up view images photographed from directly above to reveal structural details. All images were acquired under controlled studio lighting with diffused illumination against a neutral white background, using fixed camera settings to ensure consistent quality and exposure. The standardized acquisition protocol enables comprehensive feature learning by providing dual perspectives that capture both surface textures and geometric characteristics essential for robust LoRA training.

The dataset underwent rigorous quality control including sharpness verification, exposure consistency checks, and annotation accuracy validation. This balanced multi-view configuration allows LoRAs to recognize objects from multiple perspectives while capturing fine-grained differences between similar categories. The equal distribution across categories and viewpoints prevents bias during LoRA adaptation and ensures stable training for the distillation process.


\subsection{Implementation Details}

\subsubsection{Distillation Phase Configuration.}
The distillation process is structured as a two-stage training pipeline, designed to systematically transfer knowledge from the teacher experts to the unified student model. Both Stage 1 and Stage 2 follow identical training configurations. Each stage is trained for 5,000 iterations to ensure sufficient convergence of the student's features while maintaining computational efficiency. We employ Low-Rank Adaptation (LoRA) with a rank parameter set to 16. This rank provides an optimal balance between parameter efficiency and model expressiveness, allowing the adapter to capture nuanced, domain-specific features without introducing excessive parameters. This LoRA configuration allows for effective fine-tuning while reducing the number of trainable parameters compared to full fine-tuning approaches.

\subsubsection{ForgeDreamer Training Protocol.}
The ForgeDreamer training process follows a schedule spanning 5,000 iterations. The training begins with a warm-up phase covering the initial 1,500 iterations, during which the learning rate increases to its target value. This warm-up strategy helps stabilize the early stages of training and prevents potential optimization instabilities from aggressive initial learning rates. Following this stabilization period, the remaining 3,500 iterations proceed with the training protocol.

During ForgeDreamer training, we maintain a batch size of 4 throughout the entire training process. This batch size was selected to achieve maximum training efficiency by utilizing the available VRAM on the NVIDIA RTX 4090 24G GPU, while also ensuring stable gradient updates and maintaining reproducible results across different experimental runs.

\begin{figure*}[ht!]
\centering
\includegraphics[width=1.0\textwidth]{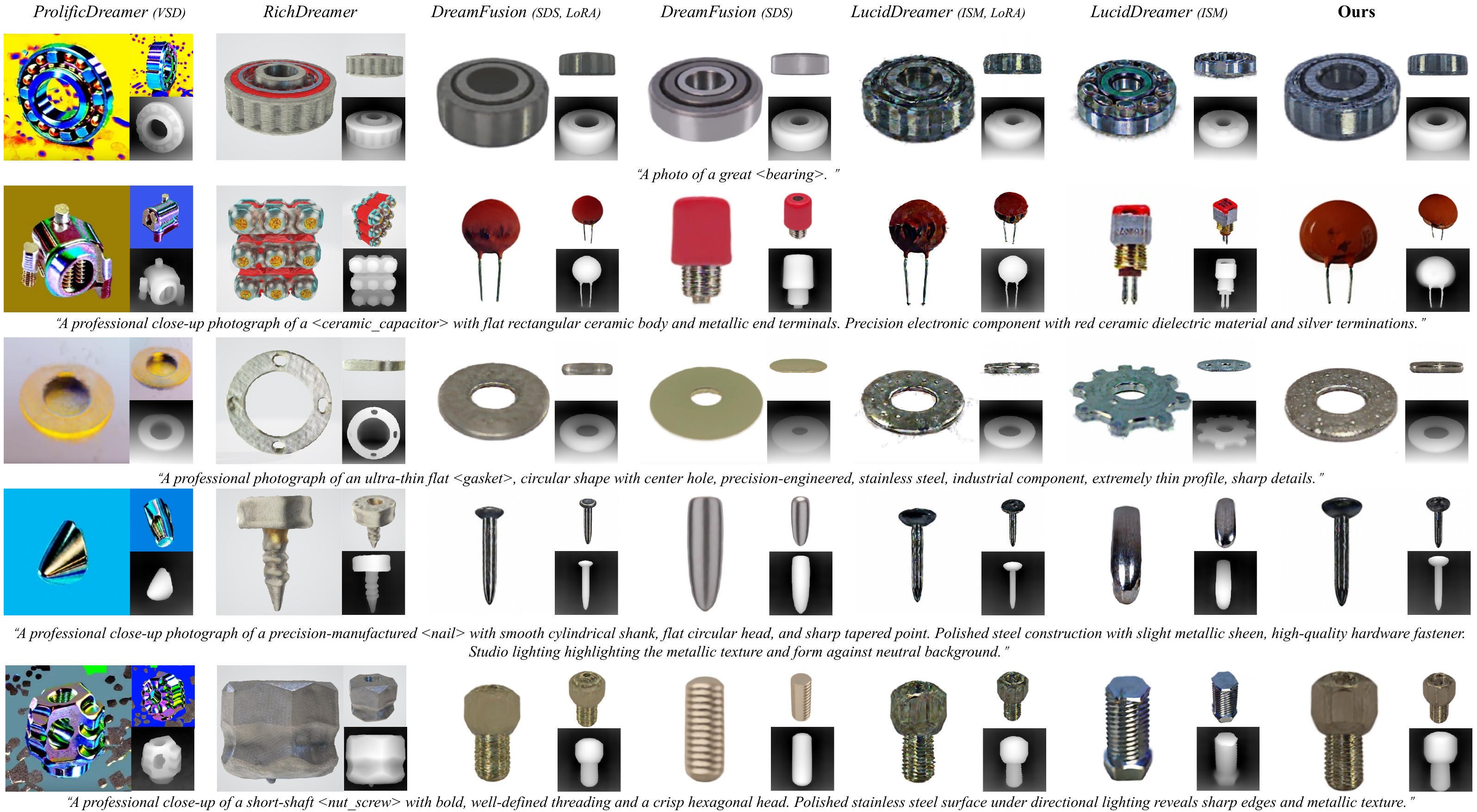} 
\caption{Qualitative Comparison with State-of-the-Art Methods. Visual results demonstrate the superior performance of our approach (remaining categories).}
\label{fig:compare_remain}
\end{figure*}

\subsubsection{Hardware and Computational Environment.}
All experimental procedures, including both the LoRA distillation phases and the complete ForgeDreamer training pipeline, are conducted on a single NVIDIA RTX 4090 24G GPU. This hardware configuration provides sufficient computational resources for the training requirements while maintaining accessibility for research purposes. The use of a single GPU setup ensures consistent experimental conditions and eliminates potential variations that might arise from distributed training configurations.

\subsection{Additional Industrial Scene Results}



We present further analysis under different industrial settings to thoroughly evaluate our method's capabilities. This includes examining the performance when processing varying numbers of industrial objects, simulating real-world scenarios that range from single-object synthesis to more complex, populated scenes. To provide a more intuitive understanding of the generation quality differences, we visualize the detailed comparative results across these different experimental configurations.

As shown in \Cref{fig:compare_remain}, which complements the visual results in the main paper, we include extended comparisons with additional baseline methods. This figure focuses on the remaining five industrial categories: \textbf{bearing, ceramic capacitor, gasket, nail, and hexagonal stud}. These visual results clearly demonstrate the robustness and generality of our approach in industrial object generation. Across these categories, our method consistently produces higher-fidelity geometry, more accurate structural details (such as the threading on studs or contacts on capacitors), and avoids common failure modes like surface oversmoothing or disconnected parts that are visible in baseline results. The comprehensive analysis reveals consistent performance improvements across these different experimental configurations within the industrial domain.

\subsection{Detailed LoRA Fusion Analysis}
Detailed analysis of the LoRA fusion mechanisms reveals significant insights into the behavior of different fusion strategies across varying configurations. Through systematic evaluation of 2, 4, and 6 LoRA configurations, we investigate both quantitative performance metrics and underlying representational changes.

\begin{figure*}[t!]
\centering
\subfloat[Cosine Similarity]{%
 \includegraphics[width=1\columnwidth]{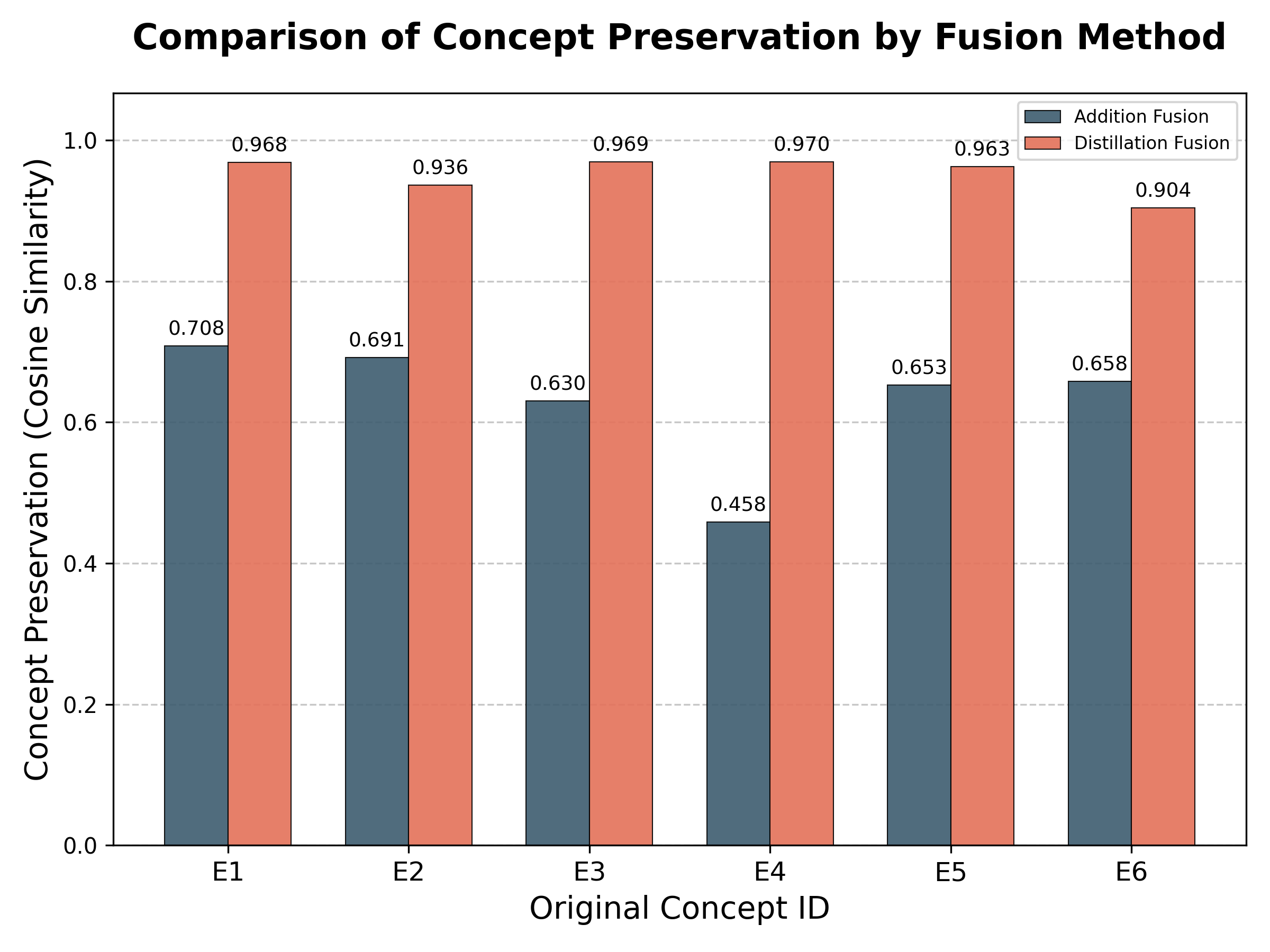}
 \label{subfig:six_cos}
}
\subfloat[PCA Analysis]{%
 \includegraphics[width=1\columnwidth]{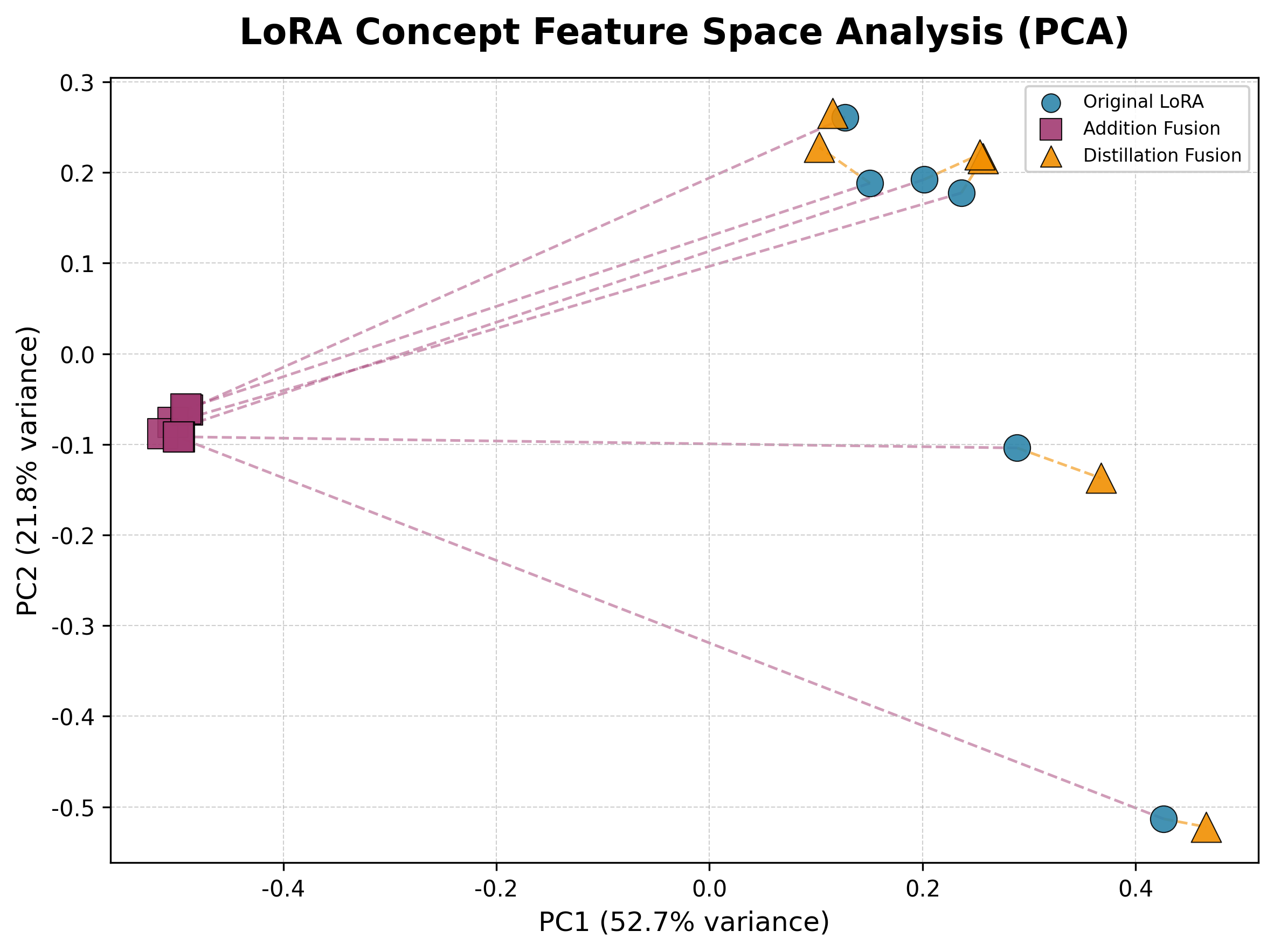}
 \label{subfig:six_pca}
}
\caption{\small Analysis of Six LoRAs Configuration: (a) Cosine similarity comparison between addition fusion and ablation fusion methods. (b) PCA visualization showing the representational space distribution and relative distances to the original LoRA.}

\label{fig:six_loras_analysis}
\centering
\subfloat[Cosine Similarity]{%
 \includegraphics[width=1\columnwidth]{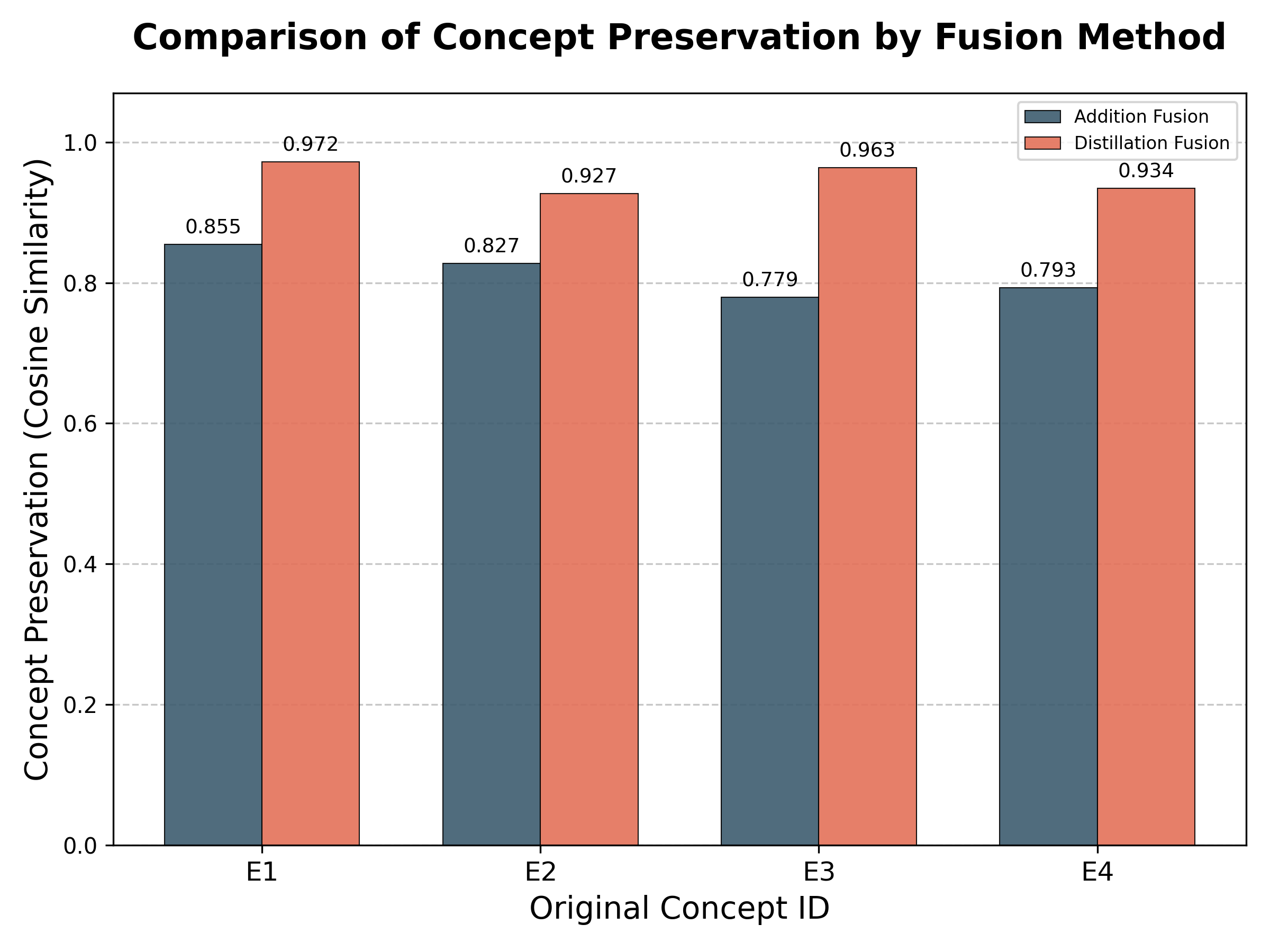}
 \label{subfig:four_cos}
}
\subfloat[PCA Analysis]{%
 \includegraphics[width=1\columnwidth]{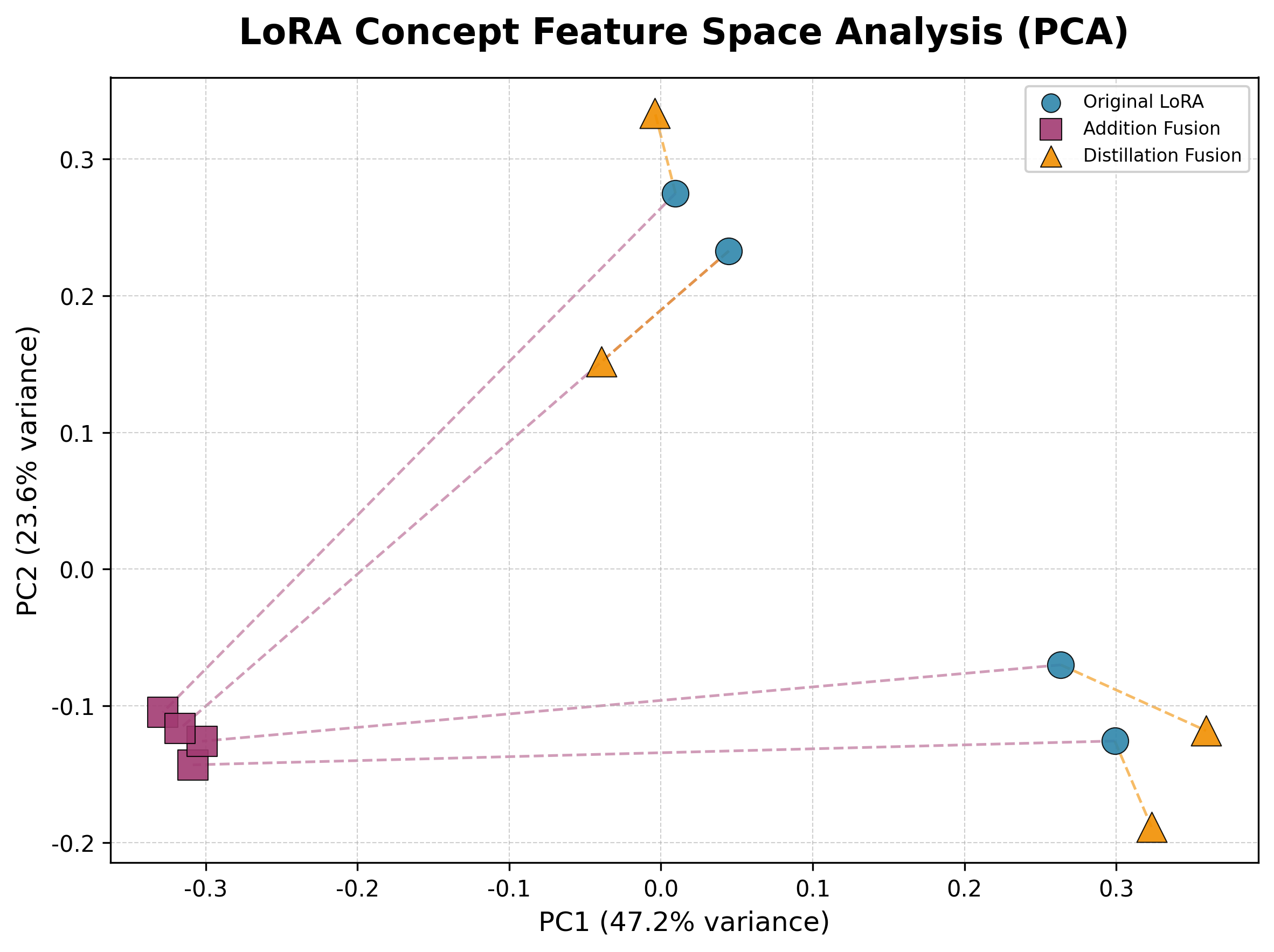}
 \label{subfig:four_pca}
}
\caption{\small Analysis of Four LoRAs Configuration: (a) Cosine similarity metrics demonstrating the effectiveness of addition fusion over ablation fusion. (b) PCA decomposition revealing the structural relationships between fused LoRAs and the original representation.}
\label{fig:four_loras_analysis}

\centering
\subfloat[Cosine Similarity]{%
 \includegraphics[width=1\columnwidth]{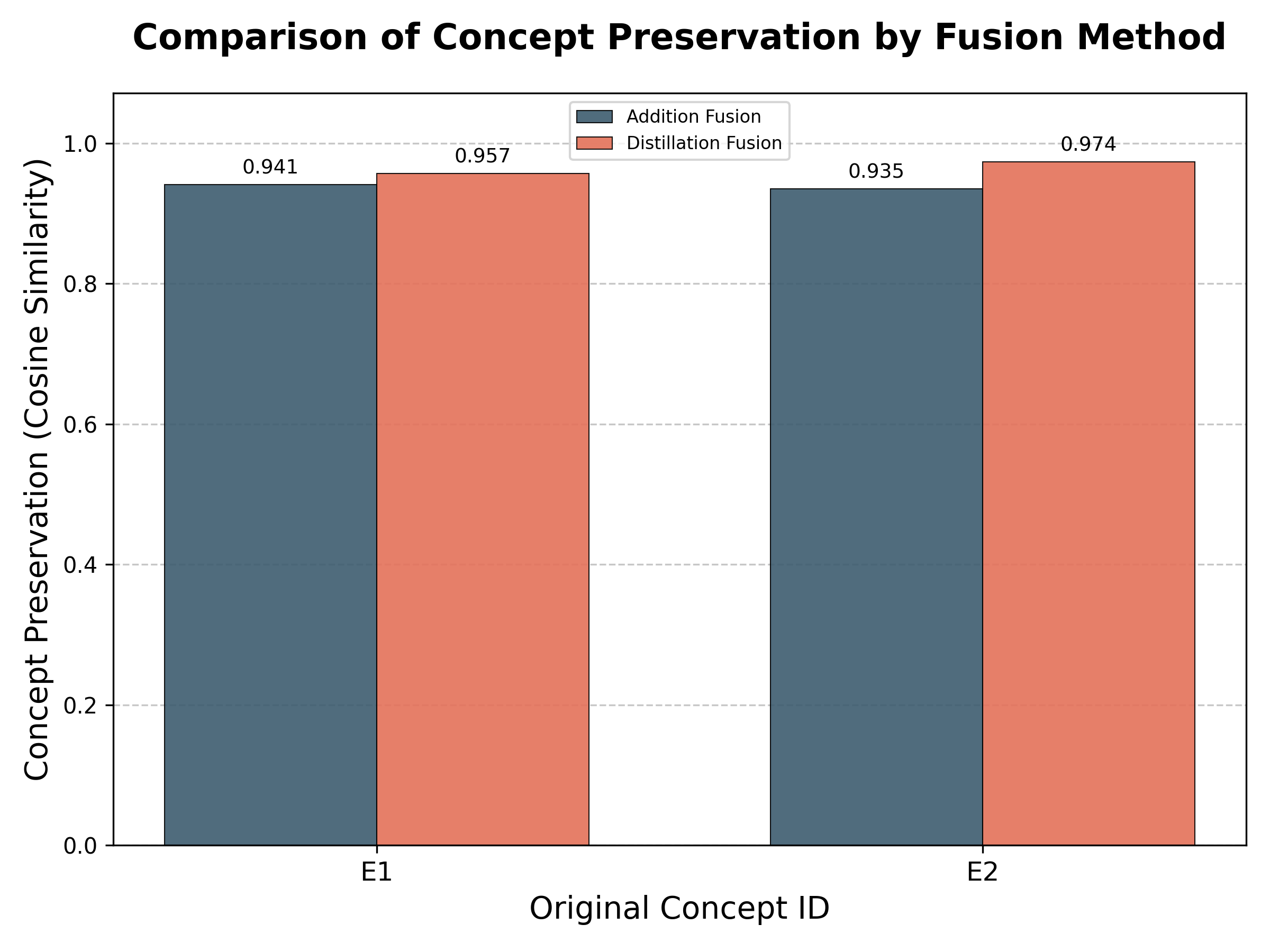}
 \label{subfig:two_cos}
}
\subfloat[PCA Analysis]{%
 \includegraphics[width=1\columnwidth]{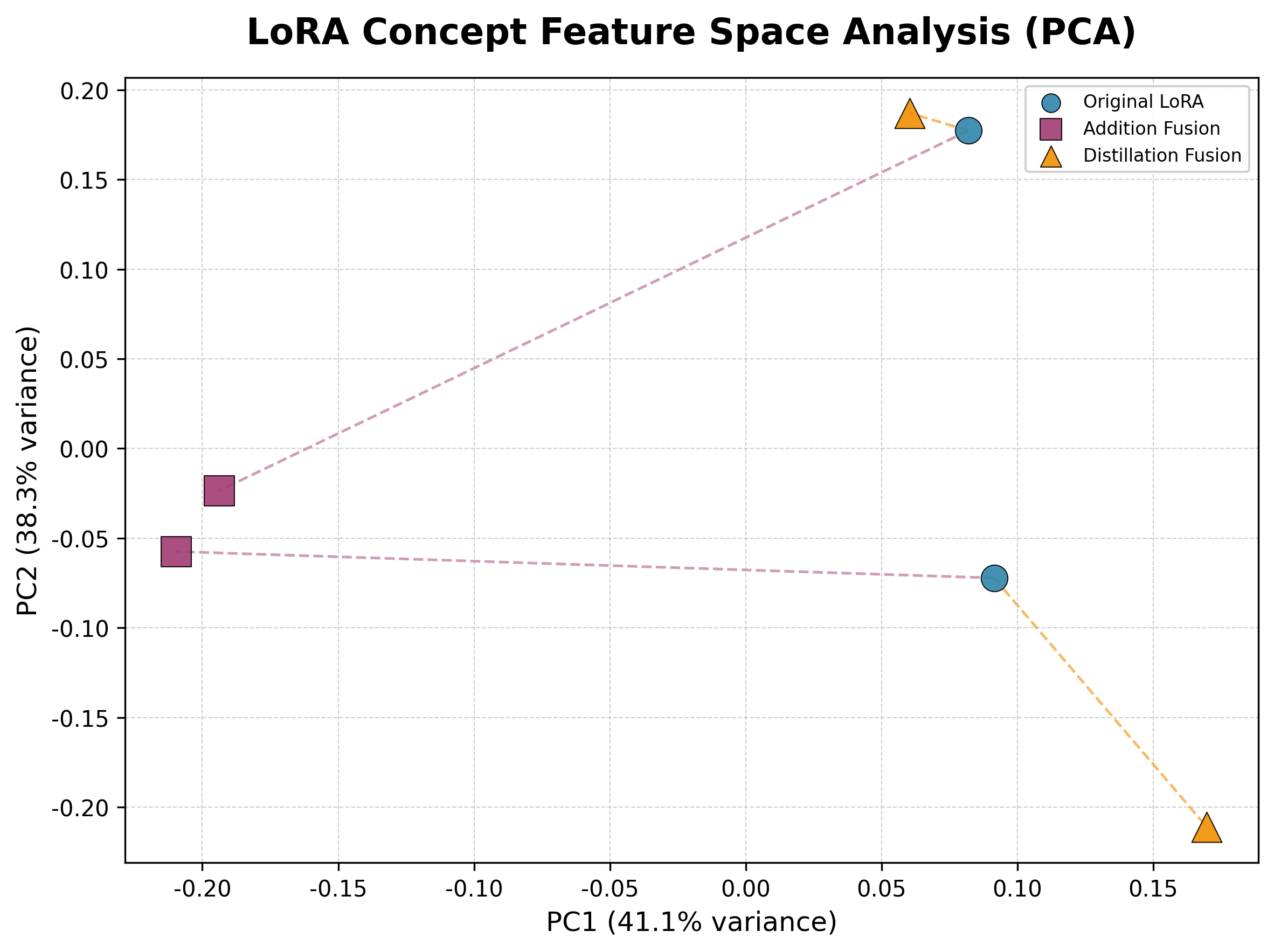}
 \label{subfig:two_pca}
}
\caption{\small Analysis of Two LoRAs Configuration: (a) Cosine similarity comparison showing the fundamental differences between fusion approaches. (b) PCA analysis illustrating the simplest case of LoRA fusion and its impact on representational geometry.}
\label{fig:two_loras_analysis}
\end{figure*}

The visualizations in \Cref{fig:six_loras_analysis}, \Cref{fig:four_loras_analysis}, and \Cref{fig:two_loras_analysis} present these analyses. The cosine similarity analysis (subplot a) provides a direct measurement of how well fused LoRAs preserve individual component characteristics, while PCA decomposition (subplot b) offers geometric insights into the structural relationships within the learned representation space.

To further quantify these findings, \Cref{tab:concept_preservation_detailed} provides a detailed comparison of concept preservation scores for our distillation method versus simple additive fusion. The data shows that our distillation method consistently and significantly outperforms additive fusion across all configurations (e.g., 0.952 vs. 0.633 average for six LoRAs), demonstrating its effectiveness in preventing catastrophic interference.

\begin{table}[t!]
\centering
\caption{Concept Preservation Scores by Fusion Method and LoRA Configuration}
\begin{tabular}{lccc}
\toprule
\textbf{Method} & \textbf{Two} & \textbf{Four} & \textbf{Six} \\
\textbf{(CLIP-ViT-L/14)} & \textbf{LoRAs} & \textbf{LoRAs} & \textbf{LoRAs} \\
\midrule
\multicolumn{4}{l}{\textit{Individual Concepts}} \\
\addlinespace[0.3em]
Emb1 (Addition) & 0.899 & 0.793 & 0.658 \\
Emb1 (Distillation) & \textbf{0.927} & \textbf{0.934} & \textbf{0.904} \\
\addlinespace[0.2em]
Emb2 (Addition) & 0.886 & 0.779 & 0.653 \\
Emb2 (Distillation) & \textbf{0.915} & \textbf{0.963} & \textbf{0.963} \\
\addlinespace[0.2em]
Emb3 (Addition) & --- & 0.855 & 0.708 \\
Emb3 (Distillation) & --- & \textbf{0.972} & \textbf{0.968} \\
\addlinespace[0.2em]
Emb4 (Addition) & --- & 0.827 & 0.691 \\
Emb4 (Distillation) & --- & \textbf{0.927} & \textbf{0.936} \\
\addlinespace[0.2em]
Emb5 (Addition) & --- & --- & 0.630 \\
Emb5 (Distillation) & --- & --- & \textbf{0.969} \\
\addlinespace[0.2em]
Emb6 (Addition) & --- & --- & 0.458 \\
Emb6 (Distillation) & --- & --- & \textbf{0.970} \\
\addlinespace[0.3em]
\midrule
\multicolumn{4}{l}{\textit{Overall Performance}} \\
\addlinespace[0.3em]
\textbf{Average (Addition)} & 0.938 & 0.814 & 0.633 \\
\textbf{Average (Distillation)} & \textbf{0.965} & \textbf{0.949} & \textbf{0.952} \\
\bottomrule
\end{tabular}
\label{tab:concept_preservation_detailed}
\end{table}

\subsection{LLM-Based Qualitative Evaluation}
To provide an objective and detailed qualitative assessment, we employed a Large Language Model (LLM) to evaluate and compare our method against the baselines. This qualitative feedback was then distilled into a quantitative ranking by tallying the LLM's preferences across all comparisons, as summarized in \Cref{tab:user_study_rank}. \Cref{tab:user_study_rank} details these results, where a lower rank indicates better user preference. Our method achieved the best possible rank (1.0) in 6 out of 10 categories for an average rank of 1.6, significantly outperforming all baselines and confirming its superior perceptual quality. The following \Cref{fig:claude_eval_a,fig:claude_eval_b,fig:claude_eval_c,fig:claude_eval_d} present the complete, verbatim responses from the LLM evaluator, which form the basis of this analysis. This evaluation provides a comprehensive and unbiased analysis of the geometry, texture, and prompt adherence of the generated results across all ten industrial categories and additional prompts.

\begin{table*}[t]
\centering
\caption{Quantitative user study ranking across ten object categories. Lower ranks are better. Best (rank 1) and second-best (rank 2) results in each column are highlighted in \textbf{bold} and \underline{underlined}, respectively.}
\label{tab:user_study_rank}
\resizebox{\textwidth}{!}{%
\begin{tabular}{l|cccccccccc|c}
\toprule
\textbf{Method} & \textbf{G. LED} & \textbf{Nut} & \textbf{R. LED} & \textbf{Resistor} & \textbf{Screw} & \textbf{Bearing} & \textbf{Capacitor} & \textbf{Hex Stud} & \textbf{Nail} & \textbf{Gasket} & \textbf{Average} \\
\midrule
ProlificDreamer~\cite{wang2023prolificdreamer} (w/o LoRA) & 7 & 7 & 7 & 7 & 7 & 7 & 7 & 7 & 7 & 6 & 6.9 \\
RichDreamer~\cite{qiu2024richdreamer} (w/o LoRA) & 6 & 6 & 6 & 6 & 6 & 4 & 6 & 6 & 4 & 3 & 5.3 \\
DreamFusion~\cite{poole2022dreamfusion} (w/ LoRA) & 3 & 5 & \underline{2} & \underline{2} & \underline{2} & \underline{2} & 4 & \underline{2} & \textbf{1} & \textbf{1} & \underline{2.4} \\
DreamFusion~\cite{poole2022dreamfusion} (w/o LoRA) & 5 & 4 & 5 & 5 & 5 & \textbf{1} & \textbf{1} & 5 & 5 & 4 & 4.0 \\
LucidDreamer~\cite{liang2024luciddreamer} (w/LoRA) & \underline{2} & 3 & 3 & 3 & 4 & 6 & 5 & 4 & 3 & 5 & 3.8 \\
LucidDreamer~\cite{liang2024luciddreamer} (w/o LoRA) & 4 & \underline{2} & 4 & 4 & 3 & 5 & \underline{2} & 3 & 6 & 7 & 4.0 \\
\midrule
Ours & \textbf{1} & \textbf{1} & \textbf{1} & \textbf{1} & \textbf{1} & 3 & 3 & \textbf{1} & \underline{2} & \underline{2} & \textbf{1.6} \\
\bottomrule
\end{tabular}%
}
\end{table*}

\subsection{Natural Scene Generation Evaluation}
To evaluate the generalizability of our approach beyond its primary industrial applications, we conducted a comparative analysis on common natural scene generation tasks. While ForgeDreamer is specifically designed and optimized for industrial object synthesis, we demonstrate that our unified LoRA fusion strategy still maintains highly competitive performance in these distinct natural scene contexts.

\begin{figure*}[t!]
\centering
\includegraphics[width=1.0\textwidth]{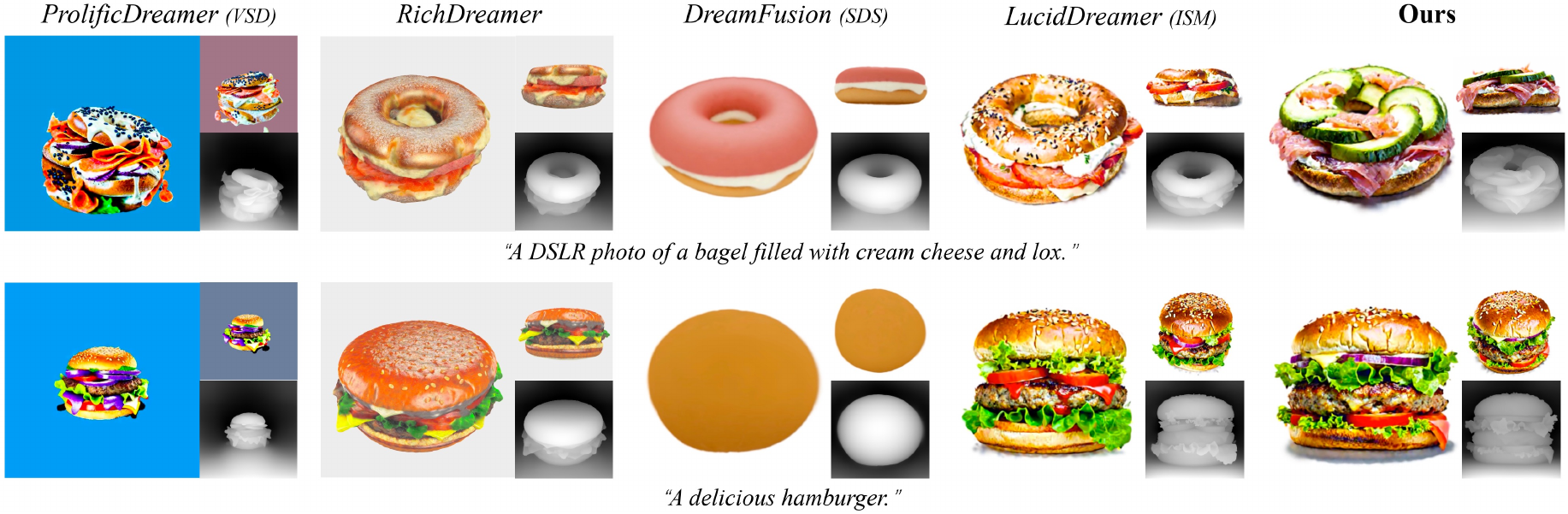} 
\caption{Qualitative comparison with state-of-the-art methods on natural scenes. Visual results demonstrate the superior performance of our approach.}
\label{fig:compare_natural}
\end{figure*}

\begin{figure*}[t]
\centering

\begin{subfigure}{0.75\linewidth}
    \includegraphics[width=\linewidth]{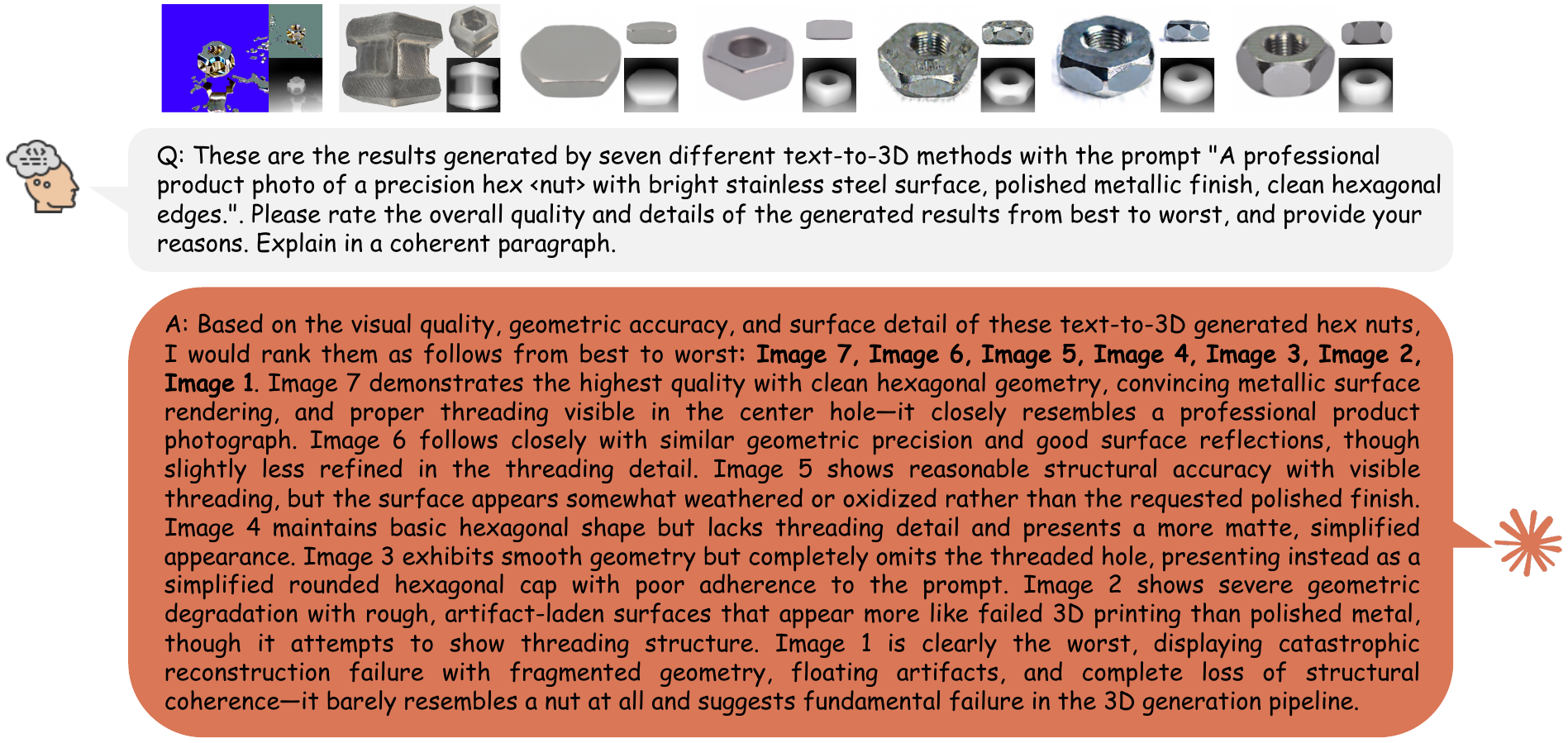}
    \caption{Nut}
\end{subfigure}

\begin{subfigure}{0.75\linewidth}
    \includegraphics[width=\linewidth]{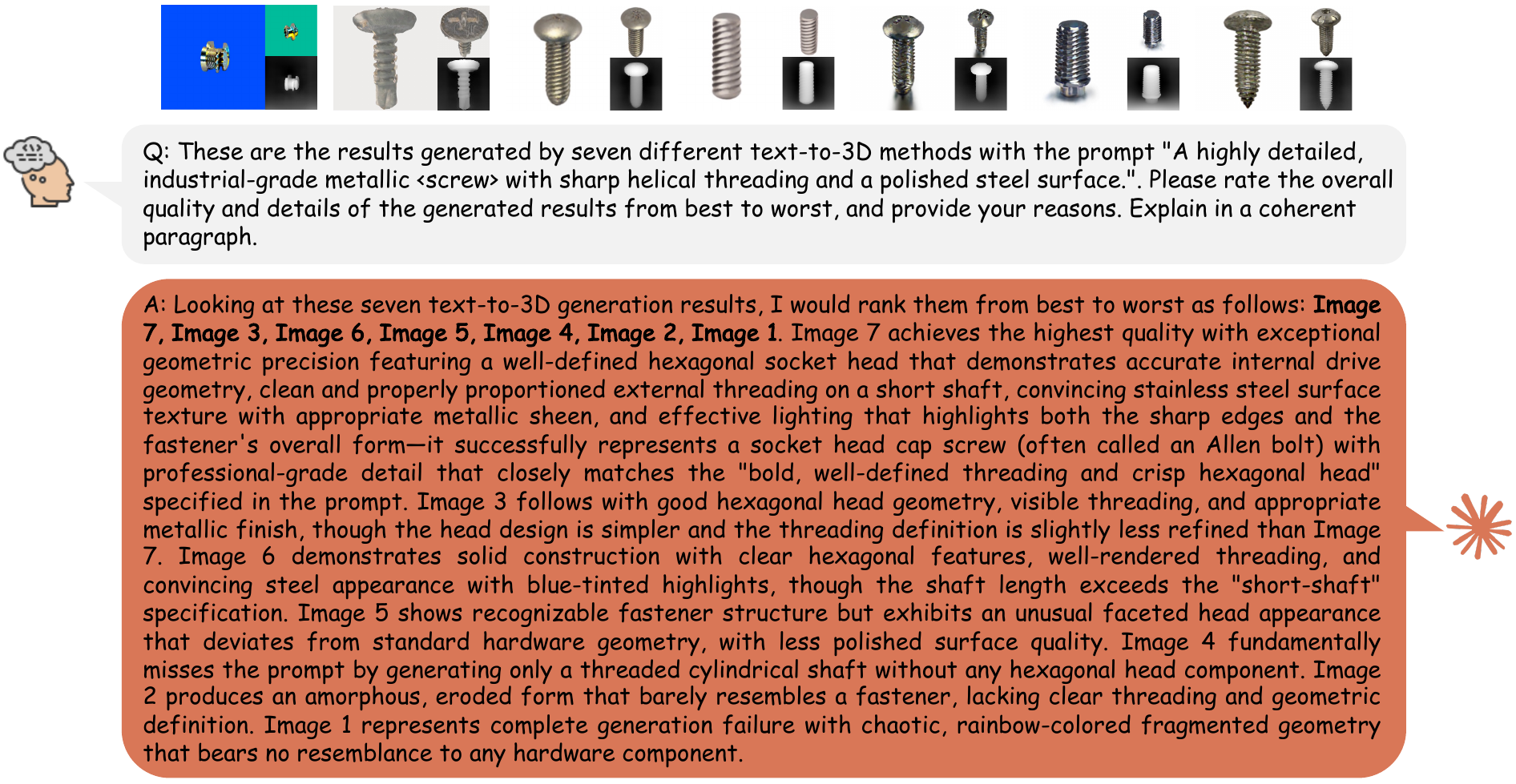}
    \caption{Screw}
\end{subfigure}

\begin{subfigure}{0.75\linewidth}
    \includegraphics[width=\linewidth]{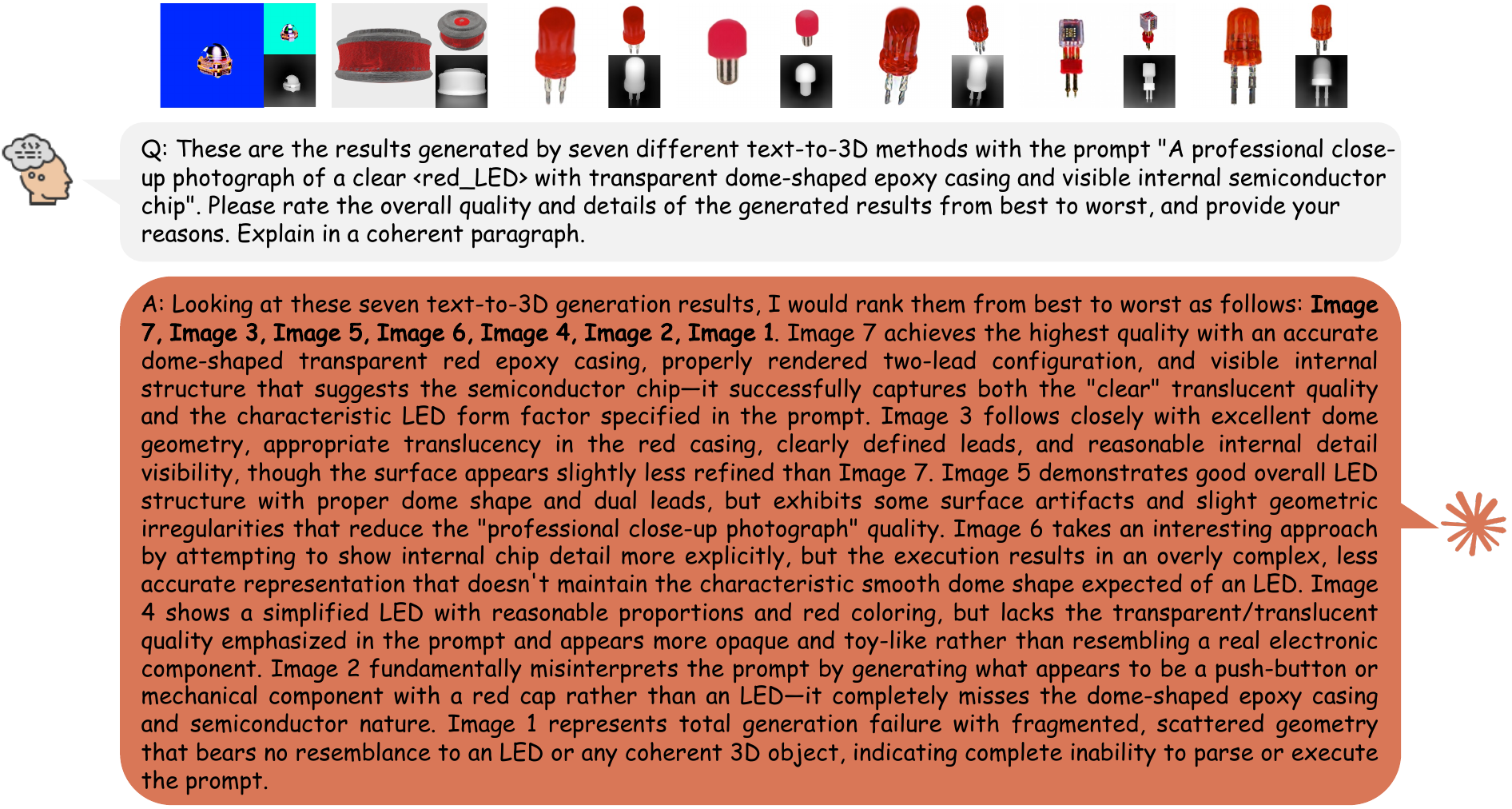}
    \caption{Red LED}
\end{subfigure}

\caption{LLM-based qualitative evaluation (Part 1-3).}
\label{fig:claude_eval_a}
\end{figure*}

\begin{figure*}[t]
\centering

\begin{subfigure}{0.75\linewidth}
    \includegraphics[width=\linewidth]{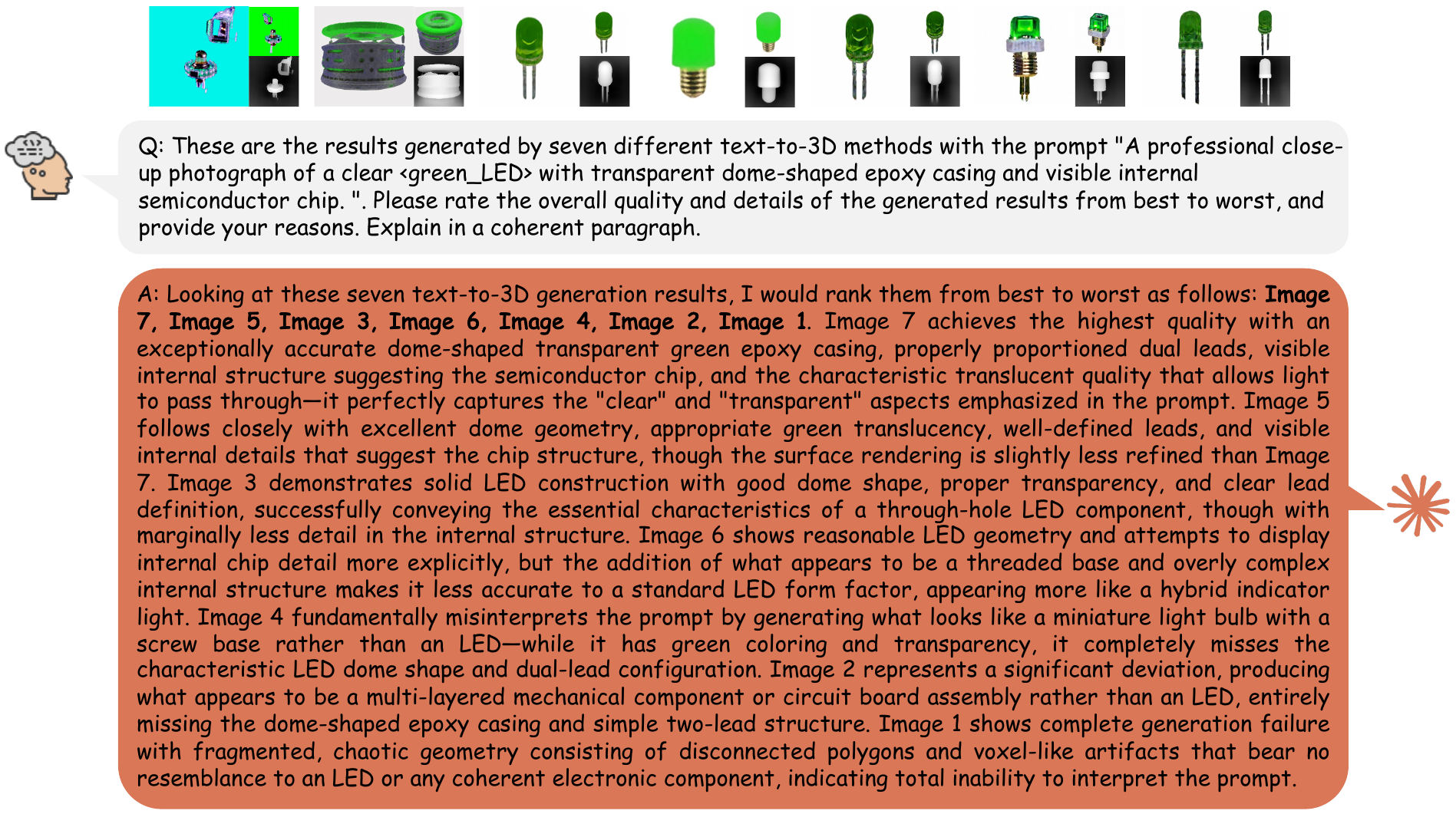}
    \caption{Green LED}
\end{subfigure}

\begin{subfigure}{0.75\linewidth}
    \includegraphics[width=\linewidth]{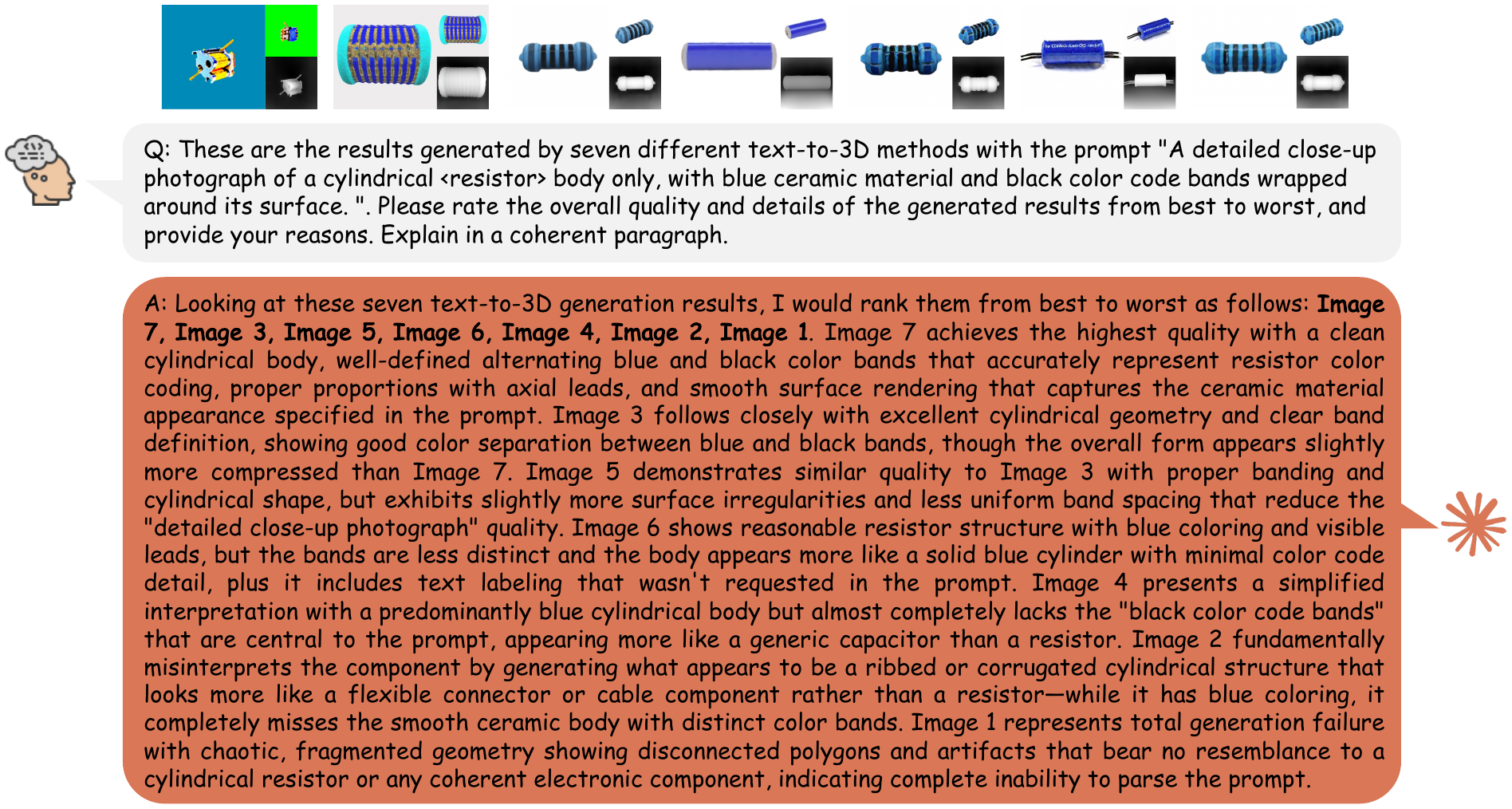}
    \caption{Resistor}
\end{subfigure}

\begin{subfigure}{0.75\linewidth}
    \includegraphics[width=\linewidth]{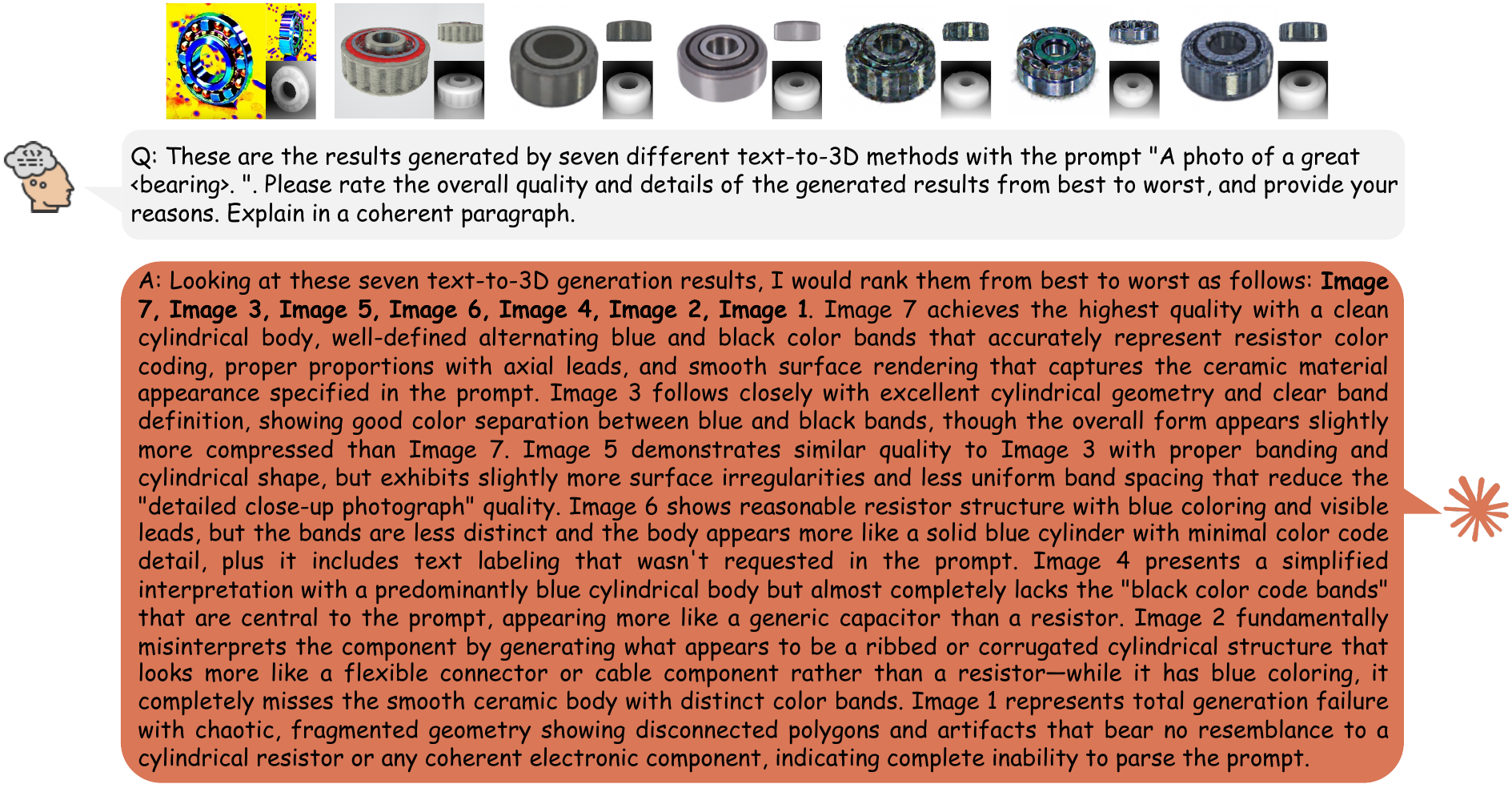}
    \caption{Bearing}
\end{subfigure}

\caption{LLM-based qualitative evaluation (Part 4-6).}
\label{fig:claude_eval_b}
\end{figure*}

\begin{figure*}[t]
\centering

\begin{subfigure}{0.75\linewidth}
    \includegraphics[width=\linewidth]{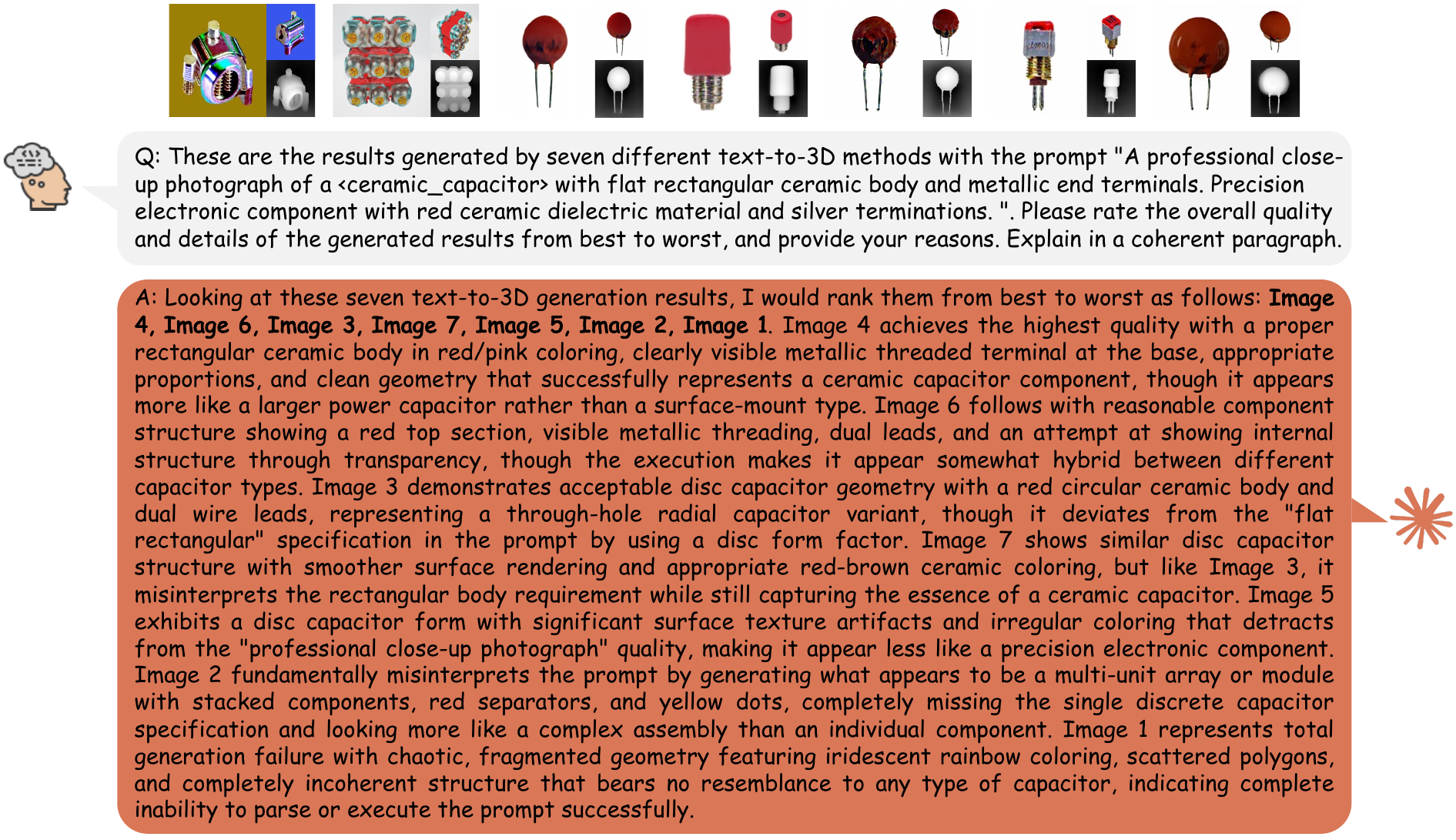}
    \caption{Ceramic Capacitor}
\end{subfigure}

\begin{subfigure}{0.75\linewidth}
    \includegraphics[width=\linewidth]{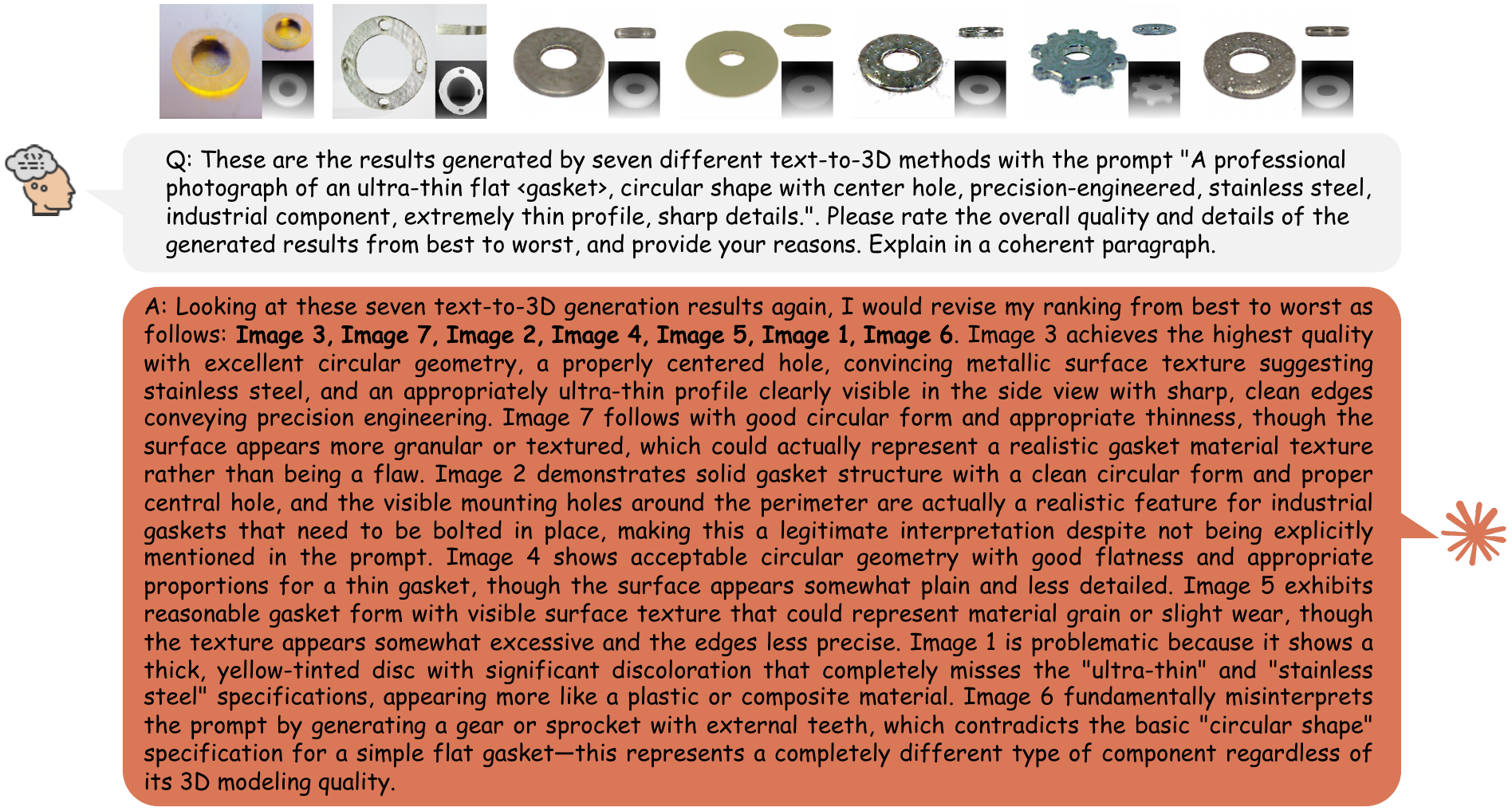}
    \caption{Gasket}
\end{subfigure}

\begin{subfigure}{0.75\linewidth}
    \includegraphics[width=\linewidth]{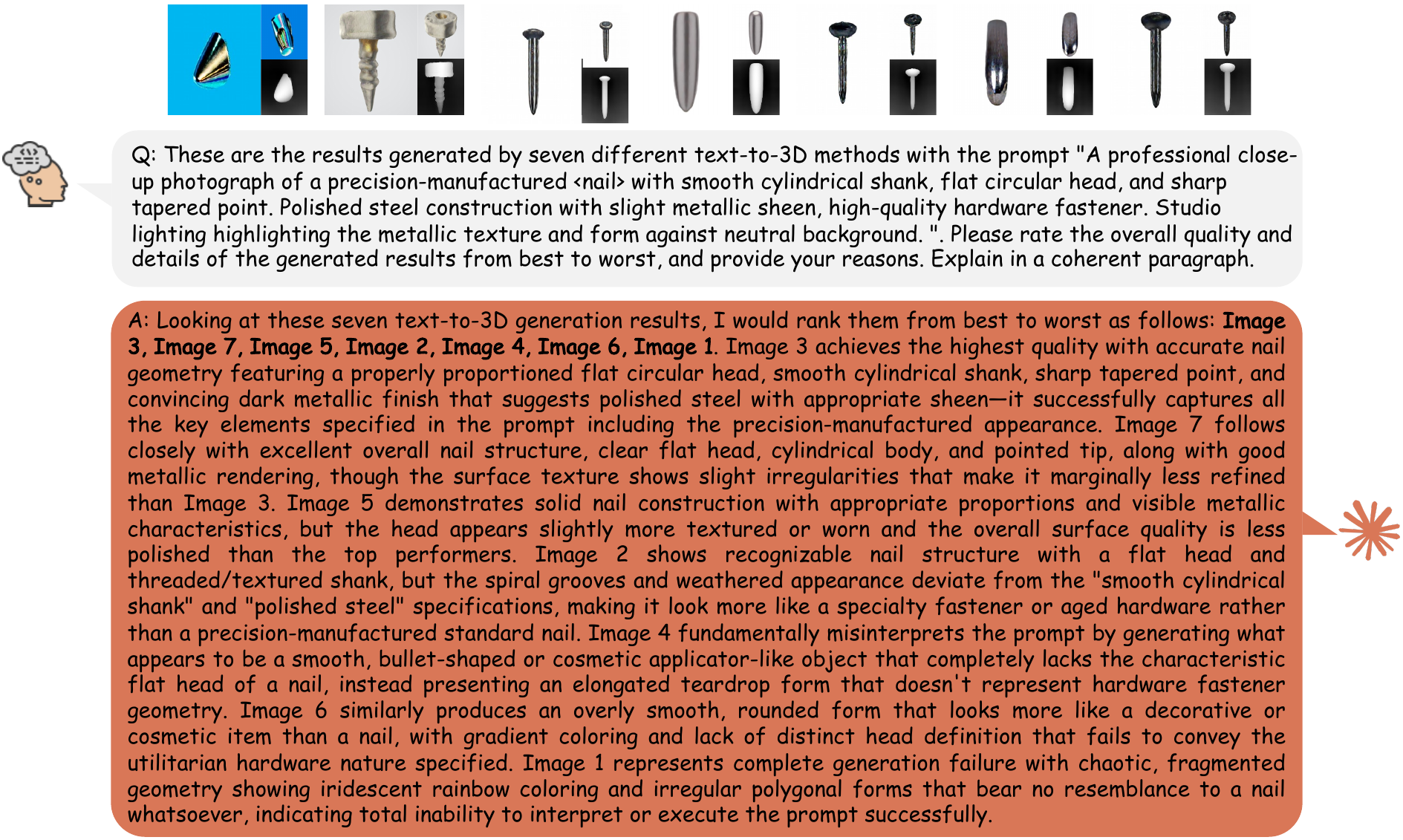}
    \caption{Nail}
\end{subfigure}

\caption{LLM-based qualitative evaluation (Part 7-9).}
\label{fig:claude_eval_c}
\end{figure*}

\begin{figure*}[t]
\centering

\begin{subfigure}{0.75\linewidth}
    \includegraphics[width=\linewidth]{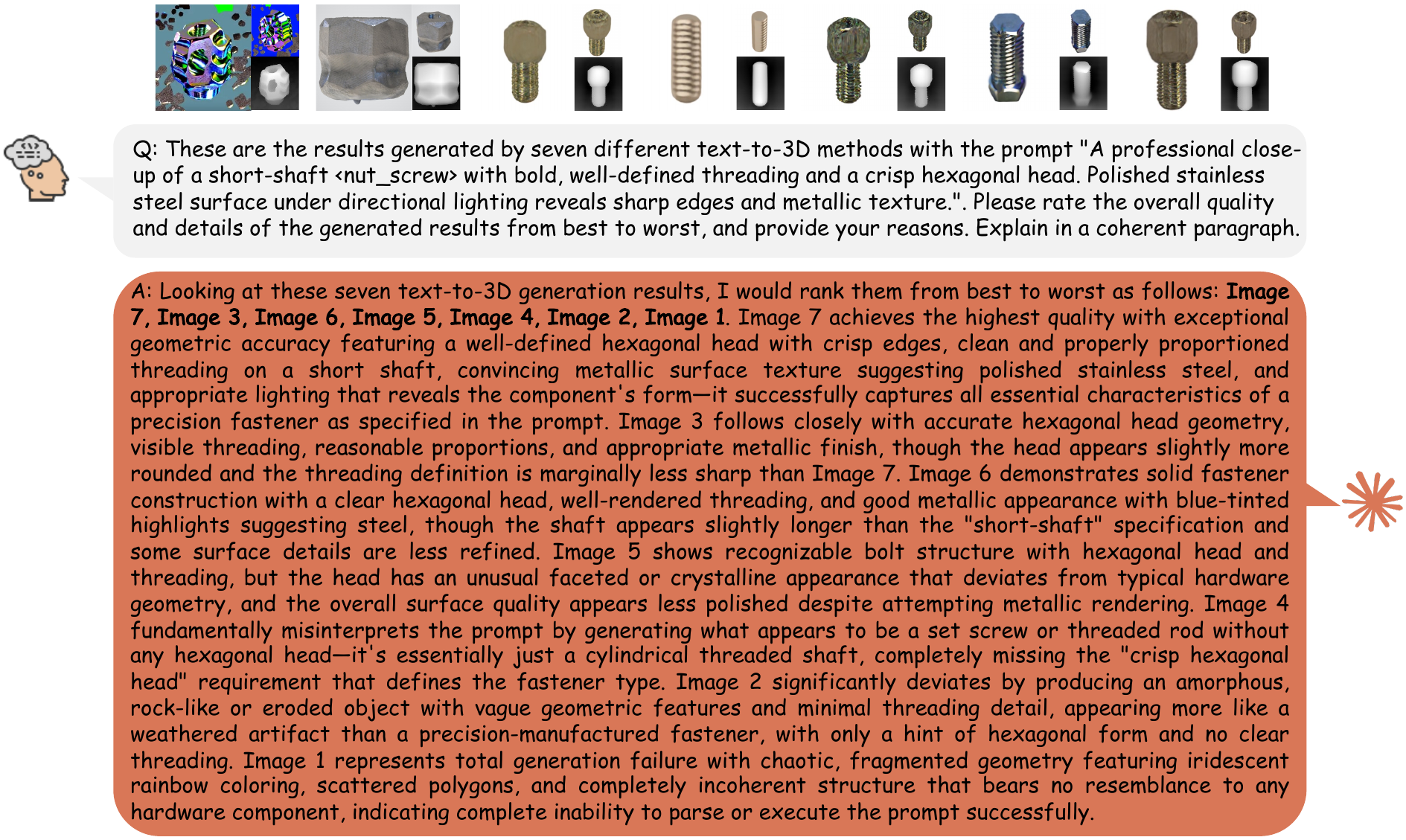}
    \caption{Hexagonal Stud}
\end{subfigure}

\begin{subfigure}{0.75\linewidth}
    \includegraphics[width=\linewidth]{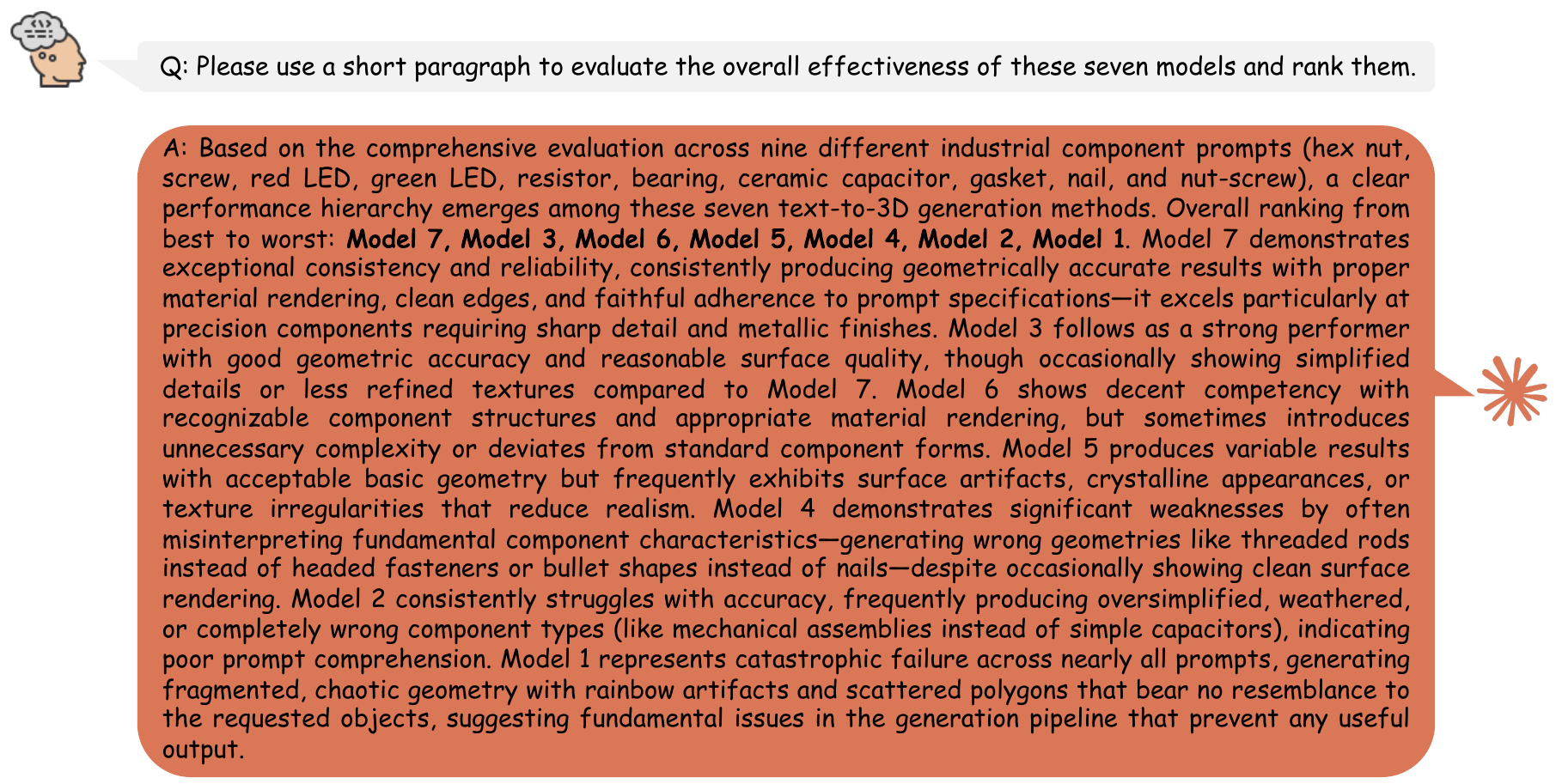}
    \caption{Overall Comment}
\end{subfigure}

\caption{LLM-based qualitative evaluation (Part 10-11).}
\label{fig:claude_eval_d}
\end{figure*}

\Cref{fig:compare_natural} presents qualitative comparisons between our method and baseline approaches on "bagel" and "hamburger" objects. The results indicate that our fusion mechanism successfully preserves the critical semantic coherence and high visual quality necessary for natural image generation. Crucially, the specialized industrial-focused training does not unduly compromise the model's ability to handle general-purpose generation tasks. This cross-domain evaluation validates that ForgeDreamer achieves its domain-specific optimization without sacrificing broader applicability, making it suitable for mixed industrial-natural generation scenarios that are commonly encountered in practical applications.

\subsection{Per-Category T3Bench Scores}
\Cref{tab:per_category_breakdown} presents the full breakdown of T3Bench quality scores across all ten industrial categories. This detailed data substantiates the findings from the main paper. It shows that our method not only achieves the highest average score (50.88) by a significant margin over the next-best competitor (47.10), but also secures a top-two result (best or second-best) in 7 out of the 10 categories.

Specifically, our method achieves the \textbf{number one} rank in four distinct categories: \textbf{G. LED}, \textbf{Screw}, \textbf{Bearing}, and \textbf{Gasket}. It also secures the \textbf{second-best} rank in three others: \textbf{R. LED}, \textbf{Hex Stud}, and \textbf{Nail}.

This demonstrates a high degree of robustness. It is particularly noteworthy that while some baselines (e.g., LucidDreamer w/o LoRA) achieve exceptionally high scores in a few specific categories where our method does not lead (such as \textbf{Nut} and \textbf{Capacitor}), they exhibit significant performance drops in others. In contrast, our method avoids catastrophic failures and maintains a high-quality baseline across the entire spectrum. This balance between high peak performance (achieving 1st place in 40\% of categories) and strong consistency (placing in the top-two 70\% of the time) is what drives the superior overall average score, confirming its robust generation quality for diverse industrial components. However, we believe this automated metric, while useful, does not fully reflect all perceptual nuances of generation quality and should be considered a strong reference rather than a complete assessment.

\begin{table*}[t!]
\centering
\caption{Quantitative comparison across ten object categories on T3Bench quality scores. Best and second-best results in each column are highlighted in \textbf{bold} and \underline{underlined}, respectively.}
\label{tab:per_category_breakdown}
\resizebox{\textwidth}{!}{%
\begin{tabular}{l|cccccccccc|c}
\toprule
\textbf{Method} & \textbf{G. LED} & \textbf{Nut} & \textbf{R. LED} & \textbf{Resistor} & \textbf{Screw} & \textbf{Bearing} & \textbf{Capacitor} & \textbf{Hex Stud} & \textbf{Nail} & \textbf{Gasket} & \textbf{Average} \\
\midrule
ProlificDreamer~\cite{wang2023prolificdreamer} (w/o LoRA) & 27.19 & 22.34 & 24.49 & 26.60 & 34.85 & 31.64 & 20.48 & 7.85 & 26.39 & 29.45 & 25.13 \\
RichDreamer~\cite{qiu2024richdreamer} (w/o LoRA) & 17.21 & 42.26 & 15.58 & \underline{42.31} & 38.33 & 21.24 & 17.83 & 24.38 & 39.90 & 23.70 & 28.27 \\
DreamFusion~\cite{poole2022dreamfusion} (w/o LoRA) & 32.62 & \underline{69.72} & 29.92 & 28.49 & 46.59 & 28.38 & \underline{35.45} & 36.66 & 52.27 & \underline{59.02} & 41.91 \\
DreamFusion~\cite{poole2022dreamfusion} (w/ LoRA) & 48.27 & 63.31 & 31.98 & 38.40 & 56.29 & 22.76 & 21.12 & 50.61 & \textbf{62.27} & 53.27 & 44.83 \\
LucidDreamer~\cite{liang2024luciddreamer} (w/o LoRA) & 44.46 & \textbf{71.37} & 31.24 & \textbf{53.23} & 53.97 & \underline{36.86} & \textbf{45.51} & \textbf{58.93} & 39.66 & 35.72 & \underline{47.10} \\
LucidDreamer~\cite{liang2024luciddreamer} (w/ LoRA) & \underline{48.45} & 64.62 & \textbf{53.98} & 30.16 & \underline{57.31} & 20.60 & 22.45 & 54.23 & 57.39 & 58.30 & 46.75 \\
\midrule
Ours & \textbf{51.68} & 65.13 & \underline{47.23} & 38.59 & \textbf{58.42} & \textbf{45.15} & 28.63 & \underline{57.12} & \underline{57.55} & \textbf{59.25} & \textbf{50.88} \\
\bottomrule
\end{tabular}%
}
\end{table*}

\subsection{Cross-Configuration Analysis.}

Comparing across all configurations (Figures~\ref{fig:six_loras_analysis}, \ref{fig:four_loras_analysis}, and \ref{fig:two_loras_analysis}), we observe several consistent patterns that demonstrate the robustness of our approach:

\begin{itemize}
\item \textbf{Scalability}: Our distillation fusion (labeled "Addition" in the charts) consistently outperforms ablation fusion across all LoRA quantities, with performance advantages becoming more pronounced as complexity increases. Our method effectively handles multi-LoRA integration without suffering from interference effects that typically plague naive fusion strategies.

\item \textbf{Stability}: The cosine similarity scores remain stable as the number of LoRAs increases, indicating robust fusion behavior. Unlike traditional methods that exhibit degradation with increased complexity, our approach maintains consistent performance even when integrating six different LoRAs.

\item \textbf{Representational Coherence}: PCA analysis shows that our fusion maintains better alignment with the original LoRA space. The geometric analysis reveals that our fusion strategy preserves essential directional characteristics while creating meaningful combinations, evidenced by tighter clustering patterns and reduced variance in principal components.
\end{itemize}

The cross-configuration analysis reveals our fusion method's performance advantage over ablation fusion becomes pronounced as the number of LoRAs increases. This suggests the benefits our approach are not merely additive, but are amplified when tackling complex, high-demand fusion scenarios. These quantitative findings, when combined with the comprehensive qualitative comparisons presented in \Cref{fig:compare_remain}, provide strong evidence for the effectiveness and the superior generalizability of our proposed LoRA fusion strategy.

\end{document}